\documentclass[10pt,twocolumn,letterpaper]{article}

\usepackage{cvpr}              % To produce the CAMERA-READY version
\usepackage{algorithm}
\usepackage{algorithmic}
\usepackage{amsmath}
\usepackage{bbm}
\usepackage{graphicx}
\usepackage{multirow}
\usepackage{threeparttable}
\usepackage{pifont}
\usepackage{bm}
\usepackage{upgreek}
\usepackage{dsfont}
\usepackage[accsupp]{axessibility}

\definecolor{cvprblue}{rgb}{0.21,0.49,0.74}

\usepackage[pagebackref,breaklinks,colorlinks,allcolors=cvprblue]{hyperref}

\def\paperID{11564} % *** Enter the Paper ID here
\def\confName{CVPR}
\def\confYear{2025}

\title{Uncertainty Meets Diversity: A Comprehensive Active Learning Framework for Indoor 3D Object Detection}

\vspace{-0.1in}
\author{Jiangyi Wang, Na Zhao\thanks{Corresponding Author: na\_zhao@sutd.edu.sg}\\
Singapore University of Technology and Design (SUTD)\\
{\tt\small wangjiangyi0519@gmail.com, na\_zhao@sutd.edu.sg}
}

\begin{document}
\maketitle
\begin{abstract}
% AL background
Active learning has emerged as a promising approach to reduce the substantial annotation burden in 3D object detection tasks, spurring several initiatives in outdoor environments. 
% difficulty of 3D indoor detection
However, its application in indoor environments remains %largely 
unexplored. 
Compared to outdoor 3D datasets, indoor datasets face significant challenges, including fewer training samples per class, a greater number of classes, more severe class imbalance, and more diverse scene types and intra-class variances.
% variability. 
% our approach
This paper presents the first study on active learning for indoor 3D object detection, where we propose a novel framework tailored
% specifically designed 
for this task. Our method incorporates two key criteria - uncertainty and diversity - to actively select the most 
ambiguous and informative unlabeled samples for annotation. The uncertainty criterion accounts for both inaccurate detections and undetected objects, ensuring that the most ambiguous samples are prioritized. 
% Meanwhile, the diversity criterion is formulated as a joint optimization problem that maximizes both scene types and intra-class variation, using a new Class-aware Adaptive Prototype (CAP) bank.
Meanwhile, the diversity criterion is formulated as a joint optimization problem that maximizes the diversity of both object class distributions and scene types, using a new Class-aware Adaptive Prototype (CAP) bank. 
The CAP bank dynamically allocates representative prototypes to each class, helping to capture %intra-class diversity for different object categories. 
varying intra-class diversity across different categories.
% result
% We evaluate the proposed method on the SUN RGB-D and ScanNetV2 datasets, comparing it with various baselines, and demonstrate that it outperforms them by a significant margin. 
We evaluate our method on SUN RGB-D and ScanNetV2, where it outperforms baselines by a significant margin, achieving over 85\% of fully-supervised performance with just 10\% of the annotation budget.
\end{abstract}     
\vspace{-2ex}
\section{Introduction}
\label{sec: intro}
\begin{figure}[t]
    \centering
    \includegraphics[width=\linewidth]{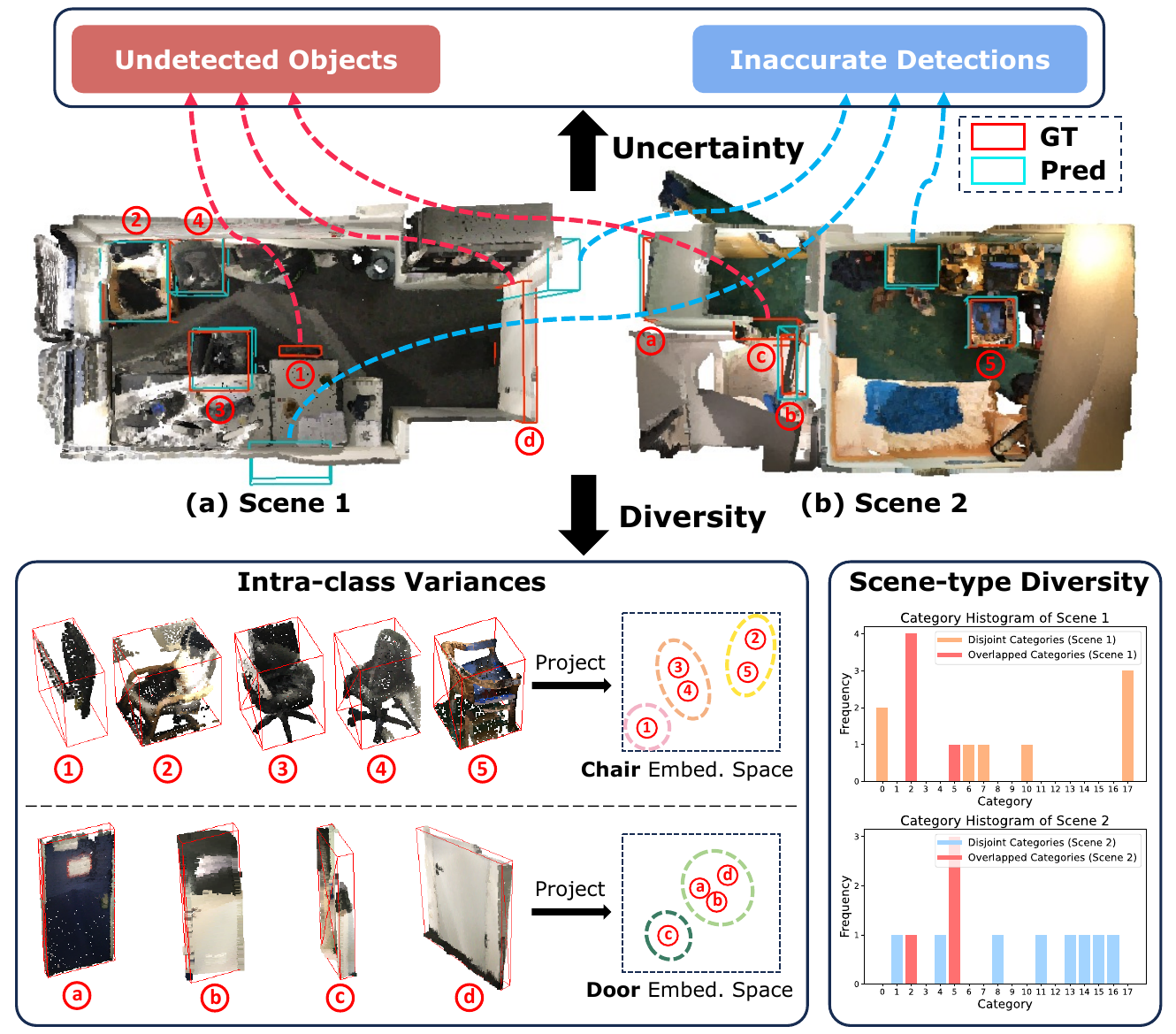}
    \caption{\small{\textbf{Challenges in active learning for indoor 3D object detection, including both uncertainty and diversity aspects.} We show two indoor scenes on ScanNetV2, with red boxes for ground truths and blue for predictions (10\% data using CAGroup3D detector), displaying only `chair' and `door' categories for clarity. For \textit{uncertainty}, the presence of \textit{undetected objects} and \textit{inaccurate detections} undermines the quality of uncertainty estimation. For \textit{diversity}, high \textit{scene-type diversity} and varying \textit{intra-class variances} in indoor environments complicate diverse sample selection. 
    This work presents a hybrid approach to address these challenges.}}
    \label{fig: teaser}
    \vspace{-3ex}
\end{figure}

%% 1. 3D object detection; 3D tasks require vast number of data; some solutions (semi-supervised, self-supervised, few-shot); why AL
3D object detection, which aims to localize and identify objects within a 3D scene, is crucial for various applications, including robotics~\cite{gashler2011temporal}, autonomous driving~\cite{li2022bevformer, yang2023bevformer}, and augmented/virtual reality~\cite{Watson_2023_CVPR}. 
While deep learning techniques have achieved significant success in 3D object detection~\cite{wang2022cagroup3d, shen2023v, misra2021end}, most existing approaches are data-hungry~\cite{shen2023v}, requiring large amounts of annotated data for training. However, annotating objects in a 3D scene is a labor-intensive process. For example, annotating a single scene in the SUN RGB-D dataset takes over \textbf{1.9 minutes} per object, with an average of approximately \textbf{14 objects} per scene~\cite{song2015sun}. 
To mitigate the high costs of 3D annotation, active learning (AL)~\cite{mi2022active, nguyen2004active, sener2018active} offers a promising solution. AL seeks to maximize the utility of limited labeling resources by actively selecting the most informative samples for human annotation. Through this process, AL can achieve performance comparable to fully-supervised methods, but with significantly fewer labeled samples. %In comparison to other data-efficient learning methods, such as semi-supervised learning, AL is often more effective at reducing the need for extensive labeled data.

Although active learning has proven effective in image-based tasks such as image classification and object detection, using uncertainty~\cite{roth2006margin, citovsky2021batch, joshi2009multi} or/and diversity~\cite{nguyen2004active, sener2018active, yang2015multi} as selection criteria, its application to 3D object detection remains largely underexplored. To date, only a few studies \cite{luo2023kecor, luo2023exploring, ghita2024activeanno3d} have applied AL techniques to outdoor 3D object detection. 
% For example, CRB~\cite{luo2023exploring} introduces three heuristic criteria to progressively select samples with concise labels, representative features, and balanced geometry. KECOR~\cite{luo2023kecor} proposes a kernel coding rate maximization method to select the most informative samples. 
For instance, CRB~\cite{luo2023exploring} uses label conciseness to capture uncertainty, along with feature representativeness and geometric balance to enhance diversity for sample selection.
KECOR~\cite{luo2023kecor} favors samples whose latent features have the highest coding length, thereby ensuring diversity.
While these methods demonstrate effectiveness in outdoor scenarios, they are suboptimal for indoor environments, as illustrated in Tab.~\ref{tab: experiment mAP}, due to the unique challenges posed by 3D object detection in indoor settings.

%%%%% HERE can link the Table~motivation %%%%%
% On the one hand, a 3D indoor dataset~\cite{dai2017scannet} could be \textbf{6.2x smaller} in scale but contain \textbf{2.3x more objects} per scene, with more severe class imbalance, compared to an outdoor dataset~\cite{geiger2013vision}, as demonstrated in~\cite{han2024dual}. 
On the one hand, a 3D indoor dataset~\cite{dai2017scannet} could be \textbf{6.2x smaller} in scale but contain \textbf{3.3x more} number of categories, with more severe class imbalance, compared to an outdoor dataset~\cite{geiger2013vision}, as shown in~\cite{han2024dual}. 
This scarcity of training samples, coupled with pronounced class imbalance, impairs 3D detection accuracy in indoor scenarios, leading to more frequent occurrences of 
% \textit{inaccurate localization}
\textit{inaccurate detection} and \textit{undetected objects}, as illustrated in Fig.~\ref{fig: teaser}. In this figure, three detections are inaccurate, incorrectly localizing background objects, while three other objects remain undetected.
These challenges underscore the need to account for both inaccurately detected proposals and undetected objects when selecting informative samples from unlabeled data.

On the other hand, unlike outdoor scenes, which primarily feature road environments with a limited set of object classes, indoor environments exhibit greater \textit{scene-type diversity} and \textit{intra-class variances}, caused by diverse layouts and variations in object shapes within each class. 
For example, as shown in Fig.~\ref{fig: teaser}, five `chair' instances form three clusters in the embedding space due to the variations in shape and occlusion.
This phenomenon in indoor environments complicate the selection of diverse samples.

In this work, we present the \textit{first solution} to apply active learning in indoor 3D object detection, utilizing both uncertainty and diversity as selection criteria. 
For \textit{uncertainty}, we design a two-pronged epistemic uncertainty estimation strategy to address \textit{inaccurate detections} and \textit{undetected objects},
% which stem from the epistemic uncertainty of the detection model. 
% Our strategy first assesses uncertainty separately for these two cases, 
% % inaccurate detections and undetected objects, 
% then integrates them into a unified 
% % uncertainty
% criterion for each sample. 
which first assesses uncertainty separately for each case, then integrates them into a unified criterion.
% Concretely, for detected proposals, we introduce an IoU-based 
% acquisition score,
% % uncertainty metric 
% % which is incorporated into the classification uncertainty measured by Shannon entropy. 
% integrated with the classification uncertainty measured by Shannon entropy. 
% Concretely, for detected proposals, we combine perturbed IoU scores with Shannon entropy to form an uncertainty score that favors ambiguously classified yet reliably localized objects.
%\textcolor{red}{Specifically, we introduce a localization-aware uncertainty score which incorporates IoU scores with Shannon entropy, capturing both classification and localization of detected proposals.}
Specifically, we introduce a localization-aware uncertainty score that combines the IoU score — measured between the original detected object and its perturbed version — with the Shannon entropy. This helps mitigate the negative effects of invalid proposals, such as background or partial detections, when estimating the classification uncertainty of detected proposals.
%Specifically, we introduce a localization-aware uncertainty score that integrates Shannon entropy with an IoU score, measured by the IoU between the original detected object and its perturbed version, to capture both the classification and localization uncertainty of each detected proposal.
To quantify the epistemic uncertainty arising from undetectability, we develop a lightweight network that estimates the number of undetected objects.
%For uncertainty, we address inaccurate detections and undetected objects by designing a two-pronged epistemic uncertainty estimation strategy. This strategy separately measures uncertainty for inaccurate detections and undetected objects, and then integrates these scores into a unified uncertainty criterion for each sample. Specifically, for detected proposals, we introduce an IoU-based uncertainty metric that decouples localization inaccuracy from classification uncertainty using an entropy measure. To quantify the epistemic uncertainty of undetected proposals, we design a lightweight network that estimates the number of undetected objects.
% For \textit{diversity}, we first introduce a Class-aware Adaptive Prototype (CAP) bank, which can dynamically allocate representative samples per class. Using the CAP Bank, we then frame the diverse sample selection problem as a joint optimization task, simultaneously optimizing scene-type and intra-class diversity.
% \textcolor{red}{For \textit{diversity}, we frame the diverse sample selection problem as an optimization task and jointly maximize \textit{scene-type diversity} and \textit{intra-class variances} on top of a novel Class-aware Adaptive Prototype (CAP) bank, which can dynamically allocate representative samples to each class.}
For \textit{diversity}, we frame the diverse sample selection problem as an optimization task that jointly maximizes \textit{scene-type diversity} and \textit{intra-class variances}, leveraging a novel Class-aware Adaptive Prototype (CAP) bank. 
% We design this CAP bank
This CAP bank is designed to capture varying intra-class variances by generating an adaptive number of representative prototypes per class. 
%Based on the observation 
Observing that different scene types exhibit distinct histograms of object categories as shown in Fig.~\ref{fig: teaser}, each composed of various
% different 
prototypes, we use the CAP bank to model both class-level and scene-level diversity.

We evaluate the proposed method on SUN RGB-D~\cite{song2015sun} and ScanNet~\cite{dai2017scannet} datasets. Using CAGroup3D~\cite{wang2022cagroup3d} as base detection model, our approach achieves 56.11\%, 61.77\% mAP@0.25 on SUN RGB-D and ScanNet, respectively, reaching 87.24\% and 85.02\% of the fully-supervised performance with only 10\% of the annotation budget.

\section{Related Work}
\label{sec: related works}

%-------------------------------------------------------------------------
%\subsection{Generic Active Learning}
\noindent
\textbf{Generic Active Learning.} 
The majority of existing works have explored active learning (AL) strategies on image classification task, which can be categorized into uncertainty-based~\cite{pmlr-v70-gal17a, wang2016cost, joshi2009multi, citovsky2021batch, roth2006margin} and diversity-based~\cite{nguyen2004active, sener2018active, yang2015multi, agarwal2020contextual, elhamifar2013convex, wu2021redal} methods based on the selection criteria used.
Uncertainty-based approaches select ambiguous samples for annotation by maximizing uncertainty scores, which can be computed in various ways, such as entropy~\cite{shannon1948mathematical, kim2021lada, pmlr-v70-gal17a}, minimum posterior probability~\cite{wang2016cost}, or 
% margin between two maximum posterior probabilities~\cite{joshi2009multi, citovsky2021batch, roth2006margin}. 
multi-class margin~\cite{joshi2009multi, citovsky2021batch, roth2006margin}.
% Shannon entropy~\cite{shannon1948mathematical} is a commonly used metric in classification that quantifies uncertainty in the 
% % predicted class distribution
% \textcolor{red}{predicted class probability}
% , while MC Dropout~\cite{pmlr-v70-gal17a} estimates uncertainty based on the predicted posterior probability, which is approximated by employing Monte Carlo Dropout. [To add: \cite{roth2006margin} description]
Shannon entropy~\cite{shannon1948mathematical} is a commonly used metric in classification that quantifies uncertainty in the probability distribution, while multi-class margin~\cite{roth2006margin} measures the uncertainty through the difference between the top two predicted probabilities.
MC Dropout~\cite{pmlr-v70-gal17a} employs Monte Carlo Dropout to approximate the posterior predicted probability and use Shannon entropy to assess the underlying uncertainty. 
%\textcolor{red}{MC Dropout~\cite{pmlr-v70-gal17a} employs Monte Carlo Dropout to approximate the predicted posterior probability.}
% MC Dropout~\cite{pmlr-v70-gal17a} approximates the posterior predicted distribution via Monte Carlo Dropout for multiple forward passes. 
Diversity-based methods~\cite{agarwal2020contextual, sener2018active, yang2015multi} select a compact subset of samples that best represent the global data distribution. For example, Core-set~\cite{sener2018active} formulates active learning as a core-set selection problem and solves it using the $k$-centroid algorithm. Recent works~\cite{ash2019deep, liu2021influence, sinha2019variational} combine both uncertainty and diversity through hybrid approaches. A representative approach, BADGE~\cite{ash2019deep}, balances uncertainty and diversity by applying the $k$-means++~\cite{wu2012advances} algorithm in the gradient space, which considers both magnitude and direction of gradient vectors.
%Diversity-based methods~\cite{agarwal2020contextual, sener2018active, yang2015multi} select a concise subset of samples to describe the global data distribution. For example, Core-set~\cite{sener2018active} formulates active learning as core-set selection and solves it with $k$-centroid algorithm. 
% CDAL~\cite{agarwal2020contextual} replaces Euclidean distance in Core-set with pairwise contextual diversity. 
%Recently works~\cite{ash2019deep, liu2021influence, sinha2019variational} take the advantages of both types via hybrid approaches. For instance, \textcolor{red}{BADGE}~\cite{ash2019deep} balances the uncertainty and diversity by applying k-means++~\cite{wu2012advances} algorithm to the gradient space, which takes both magnitude and direction of the gradient vectors into consideration.

\begin{figure*}
    \centering
    \includegraphics[width=.95\linewidth]{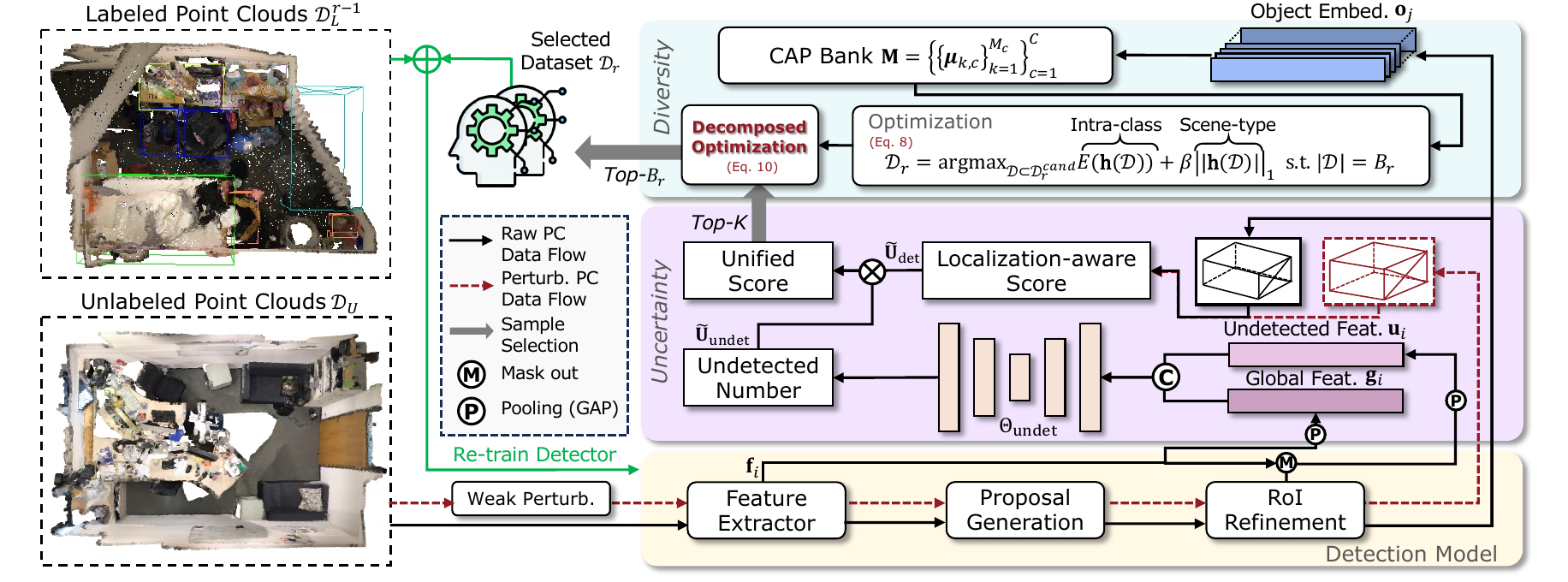}
    %\caption{An illustration of the workflow of the proposed active learning strategy for 3D detection.}
    \caption{\small{
    %\textbf{Framework overview of our proposed active learning (AL) strategy for indoor 3D object detection}.
    \textbf{Overview of our proposed AL framework for indoor 3D object detection, exploiting both uncertainty and diversity}. 
    In the $r$-th round, we first construct a candidate pool of size $\delta\cdot B_r$ using the Top-$K$ unified uncertainty score, which accounts for both inaccurate detections and undetections. Then, we jointly optimize the intra-class and scene-type diversity to select the $r$-th selected dataset $\mathcal{D}_r$. Lastly, along with the previous labeled point clouds $\mathcal{D}_{L}^{r-1}$, we retrain the model until %the size of labeled point clouds reaches budget $B$.
    the total labeled dataset reaches the annotation budget $B$.}}
    \label{fig: architecture}
    \vspace{-0.1in}
\end{figure*}

%-------------------------------------------------------------------------
%\subsection{3D Object Detection}
\vspace{0.01in}
\noindent
\textbf{3D Object Detection.}
We focus on indoor 3D object detection and exclude outdoor methods~\cite{arnold2019survey}, which differ significantly.
%Since our focus is on indoor 3D object detection, we exclude prior works on outdoor 3D detection, which differ significantly from indoor methods. For a comprehensive survey on outdoor 3D detection, we refer readers to \cite{arnold2019survey}. 
Prior indoor 3D object detection methods can be categorized into two lines in terms of underlying representation, \ie, voxel-based methods~\cite{shen2020frustum, maturana2015voxnet, gwak2020generative}, and point-based methods~\cite{zhou2022point, wang2024fly, qi2019deep, xie2020mlcvnet, xie2021venet}. Voxel-based methods convert point cloud to voxel grids and use 3D convolution networks to address the irregularity of point cloud. In contrast, based on point-level feature, point-based methods use a voting mechanism~\cite{qi2019deep, xie2021venet} to reduce the search space and encourage points to vote towards their corresponding object centers, followed by grouping the vote points into clusters and then generate object proposals. In this work, we adopt the state-of-the-art method CAGroup3D~\cite{wang2022cagroup3d} in indoor 3D object detection, 
% which combines the advantages of the two lines of work and designs a two-stage fully sparse convolutional 3D detection framework. It extracts voxel-wise features from raw point clouds and leverages `voting and grouping' mechanism.
taking the advantages of the two lines of work through a two-stage fully sparse convolutional 3D detection framework. It extracts voxel-wise features from raw point clouds and leverages `voting and grouping' mechanism to generate object proposals.

\noindent
\textbf{Active Learning for Object Detection.}
Despite extensive studies on AL in image classification, applying it to image-based object detection is challenging due to the task's complexity, which involves both localization and classification of multiple objects. Several works~\cite{kye2023tidal, mi2022active, wu2022entropy, yuan2021multiple, yang2024plug, agarwal2020contextual} have investigated active object detection by adapting generic AL techniques to this task.  
% , employing methods such as [describe types of methods]. 
For example, CDAL~\cite{agarwal2020contextual} designs contextual diversity to capture spatial and semantic context diversity within a dataset, while DivProto~\cite{wu2022entropy} and PPAL~\cite{yang2024plug} incorporate both diversity and uncertainty. 
%DivProto~\cite{wu2022entropy} and PPAL~\cite{yang2024plug} apply hybrid approaches to this task. 
% DivProto filters out similar instances during uncertainty estimation and uses prototypes~\cite{snell2017prototypical, xu2020cross} to ensure instance-level diversity.
% PPAL favors challenging categories via a difficulty coefficient and employs instance matching to develop a similarity measure for diverse sample selection.
% uses category-wise coefficient to refine the uncertainty estimation, and computes the similarities of multi-instance images to select diverse samples.
While these methods yield promising results in 2D object detection, they cannot be directly applied to 3D object detection due to the significant modality gap. Recently, driven by the need to reduce annotation burden in 3D object detection for autonomous driving, a few studies~\cite{luo2023kecor, luo2023exploring, ghita2024activeanno3d, feng2019deep, meyer2019automotive, moses2022localization, maisano2018reducing, lin2024exploring} have explored active learning for 3D object detection in outdoor environments. 
% Notable works include [describe a few representative works]. 
Notably, CRB~\cite{luo2023exploring} selects active samples with concise labels, representative features, and balanced density progressively. 
KECOR~\cite{luo2023kecor} leverages kernel coding rate~\cite{ma2007segmentation} to measure the richness of latent features. 
However, these outdoor methods are suboptimal for indoor environments, where 3D object detection faces unique challenges. To address these, this paper presents the \textit{first solution} for applying active learning to indoor 3D object detection, overcoming these challenges by incorporating designs to estimate uncertainty in both inaccurate detections and undetections, while simultaneously optimizing diversity with respect to scene-type and intra-class variances.

\section{Methodology}
\label{sec: method}

% We first depict the overall workflow of our proposed active learning (AL) approach in Fig.~\ref{fig: architecture}. 
% Then, to better illustrate the framework, we formulate the problem definition of LiDAR-based active object detection and describe the selected detection model.
% Finally, we introduce our AL strategy, detailing the uncertainty and diversity branches.

%-------------------------------------------------------------------------
%\subsection{Problem Definition}
% \noindent
% \textbf{Problem Definition.}
Given a point cloud $\mathcal{P}_i=\{(x,y,z)\}$, a 3D object detection model aims to predict the locations and class labels for a set of objects $\mathfrak{B}_i=\{(\mathbf{b}_k, y_k)\}$, where $\mathbf{b}_k\in \mathbb{R}^7$ represents the parameter of the $k$-th bounding box, including the center position, box size and yaw angle, 
%including the center position $(c_x, c_y, c_z)$, box size $(l, w, h)$ and yaw angle $\theta$,
and $y_k$ indicates its predicted class label.
%Given an unordered point clouds $\mathcal{P}_i=\{(x,y,z)\}$, 3D object detection is to localize and classify a set of bounding boxes $\mathfrak{B}_i=\{(\mathbf{b}_k, y_k)\}$, where $\mathbf{b}_k\in \mathbb{R}^7$ is the box annotation with center position $(c_x, c_y, c_z)$, box size $(l, w, h)$ and yaw angle $\theta$, and $y_k$ is the box label. 
In the active learning setting, our objective is to train the 3D detection model by constructing a small labeled dataset $\mathcal{D}_L$, which starts with a small portion of random samples $\mathcal{D}_L^0$
%($B_0$)
and is progressively expanded over $R$ rounds from a large pool $\mathcal{D}$.
%In the active learning setting, we aim to build a small labeled dataset $\mathcal{D}_L$ which is initially empty and expanded in $R$ rounds, from a large pool $\mathcal{D}$.
%\zn{In each active learning round $r\in \{1,..,R\}$, $B_r$ samples are selected from the unlabeled dataset $\mathcal{D}_{U}=\mathcal{D}/\mathcal{D}^{L}_r$ by an active learning policy, and labeled by an oracle, which constitutes the $r$-th selected dataset $\mathcal{D}_r$. Then, $\mathcal{D}^L$ will be updated as $\mathcal{D}^{L} \leftarrow \mathcal{D}^{L}_{r-1} \cup \mathcal{D}_{r}$ until the size of $\mathcal{D}_{L}$ reaches the final budget $B$, $\ie$, $\sum_{r=1}^{R}B_r=B$ where $B \ll |\mathcal{D}|$.}
In each active learning round %$r\in [R]$
$r\in \{1,..,R\}$, $B_r$ samples are selected from the unlabeled dataset $\mathcal{D}_{U}=\mathcal{D}/\mathcal{D}_{L}^{r-1}$ by an active learning policy, and labeled by an oracle, which constitute the $r$-th selected dataset $\mathcal{D}_r$. 
% Then, $\mathcal{D}_L$ will be updated as $\mathcal{D}_{L}^r \leftarrow \mathcal{D}_{L}^{r-1} \cup \mathcal{D}_{r}$ until the size of $\mathcal{D}_{L}$ reaches the final budget $B$, \ie, $|\mathcal{D}_L^0|+\sum_{r=1}^{R}B_r=B$ where $B \ll |\mathcal{D}|$.
Then, $\mathcal{D}_L$ will be updated as $\mathcal{D}_{L}^r \leftarrow \mathcal{D}_{L}^{r-1} \cup \mathcal{D}_{r}$ until the size of $\mathcal{D}_L$ reaches the final budget $B$, \ie, $|\mathcal{D}_L^0|+\sum_{r=1}^{R}B_r=B$ where $B \ll |\mathcal{D}|$.

%\subsection{Framework Overview}
% \noindent
% \textbf{Framework Overview.}
To tackle active 3D object detection in indoor scenarios, we propose a novel framework, illustrated in Fig.~\ref{fig: architecture}, that leverage both uncertainty and diversity as selection criteria. Building on previous work~\cite{wu2022entropy, yang2024plug, lin2024exploring} that considers both uncertainty and diversity, we adopt a budget expansion ratio $\delta \in \mathbb{Z}^+$ to first select a candidate pool $\mathcal{D}_r^{\text{cand}}$ of size $K=\delta \cdot B_r$ based on the uncertainty criterion for $r$-th active learning round. Then, from this pool, the $\mathcal{D}_r$ of size $B_r$ is chosen based on the diversity criterion. 
% we can add one sentence to introduce our framework is agnostic to the detection model. Our focus is on D_L selection. 
For the uncertainty criterion, we propose a two-pronged epistemic uncertainty estimation strategy that generates a unified score, accounting for uncertainty in both detected and undetected objects. This is achieved through a localization-aware uncertainty score and an estimate of the undetected object count (Sec.~\ref{sec: epistemic uncertainty calibration}). 
For the diversity criterion, 
%we develop a Class-aware Adaptive Prototype (CAP) bank, 
%\textcolor{pink}{which is refined from an initial coarse version, as shown in Figure~\ref{fig: architecture}}. Using the CAP bank, 
we frame the diverse sample selection as an optimization problem and propose an efficient solution that jointly optimizes intra-class variances and scene-type diversity (Sec.~\ref{sec: prototype-based diversity optimization}).

\subsection{Epistemic Uncertainty Estimation}
\label{sec: epistemic uncertainty calibration}
% We propose the novel false prediction-based uncertainty refinement scheme to address inaccurate localization in point cloud-level uncertainty estimation. Specifically, we introduce False Positive Examination (FPE) module to serve as a soft filter for those detected background bounding boxes, and False Negative Calibration (FNC) module to compensate for the undetected uncertainty.

\begin{table}[t]
    \begin{center}
    \resizebox{\linewidth}{!}{
    \begin{tabular}{c||cc||cc}
    \hline\hline
        Dataset & mAP@.25 & mAP@.50 & mAR@.25 & mAR@.50 \\
           \hline\hline
        SUN RGB-D~\cite{song2015sun}  & 28.91 & 12.48 & 67.12 & 26.09  \\
           \hline
        ScanNetV2~\cite{dai2017scannet}  & 35.75 &  20.40 & 76.65 & 42.48  \\
    \hline\hline
    \end{tabular}}
    \end{center}
    \vspace{-3ex}
    \caption{\small{\textbf{Motivation for designing epistemic uncertainty estimation.} We report the mean average precision (mAP) and mean average recall (mAR) at two IoU thresholds (0.25 and 0.50) %using two different IoU thresholds, 0.25 and 0.50
    for the initial detection model, which is trained using 2\% labeled data.}}
    \label{tab: motivation of uncertainty calibration}
    \vspace{-3ex}
\end{table}

In Tab.~\ref{tab: motivation of uncertainty calibration}, we evaluate the detection performance of a 3D object detection model trained on the
% \textcolor{red}{2\%} % a few 
initial random samples $D_L^0$ (2\% labeled data). % before applying active learning. 
As shown, the model achieves 20.40\% mAP@.50 and 42.48\% mAR@.50 on ScanNet~\cite{dai2017scannet}, indicating that approximately 80\% of the predictions are inaccurate and 60\% of ground-truth objects remain undetected. The results are even worse on the more challenging SUN RGB-D~\cite{song2015sun} dataset. These findings highlight the need to address the two types of epistemic uncertainty in 3D object detection: \textit{inaccurate detections} and \textit{undetected objects}. 
Given that the detection task involves both localization and classification, we estimate the epistemic uncertainty of inaccurate detections by considering both aspects. Specifically, we introduce a localization-aware uncertainty score that exploits an IoU score for localization reliability to combine with Shannon entropy. %for classification uncertainty. 
To handle undetections, we design a lightweight network to estimate the number of undetected objects, which serves as the uncertainty measure for undetections. These two uncertainty scores are then integrated into a unified uncertainty score, which is used for Top-$K$ selection to generate the candidate pool $\mathcal{D}_{r}^{\text{cand}}$.

\noindent
% \textbf{IoU-based Uncertainty Refinement.}
\textbf{Localization-aware Uncertainty Score.}
To estimate the uncertainty of detected objects from a point cloud $\mathcal{P}_i$, %which may arise from both classification and localization aspects, 
we first compute the classification uncertainty using the Shannon entropy~\cite{shannon1948mathematical} of the predicted class probabilities for all detected object proposals. For the $j$-th proposal, with predicted class probabilities over $C$ classes denoted as $\mathbf{p}_j=\{p_{j,c}\}_{c=1}^C \in \mathbb{R}^C$, the Shannon entropy is calculated as:
\begin{equation}
    \label{equ: instance entropy}
    E(\mathbf{p}_j) = -\sum_{c=1}^C p_{j,c} \log(p_{j,c}).
\end{equation}
However, not all detected proposals are valid, as some may correspond to background or partial objects. To mitigate the impact of these invalid proposals, we incorporate an IoU score that measures the similarity between the original detection and its perturbed version, obtained by applying weak augmentations (\eg, random rotation and scaling) to the input point cloud. The intuition is that if the model is confident about an object, small perturbations should not significantly affect its predicted position, while background or partial proposals will likely produce inconsistent predictions.
%Next, to estimate the uncertainty in localization, we compute an Intersection over Union (IoU) score by measuring the IoU difference between the original detected object and its perturbed version. The perturbed object is obtained by applying weak augmentations (\eg, random rotation and scaling) to the input point cloud $\mathcal{P}_i$ before passing it through the detection model. The underlying idea is that if the model is confident about the predicted location of an object, it should produce the same predicted position even in the presence of small perturbations.
Consequently, the localization-aware uncertainty score for detected objects in $\mathcal{P}_i$ is defined as:
\begin{equation}
    \label{equ: point cloud entropy}
    \mathbf{U}_{\text{det}}(\mathcal{P}_i):= \dfrac{\sum_{j=1}^{|\mathfrak{B}_i|}{\text{IoU}_j \cdot E(\mathbf{p}_j)}}{\sum_{j=1}^{|\mathfrak{B}_i|}{\text{IoU}_j}}.
\end{equation}
Our IoU scores serve as soft weights, adaptively reducing the contribution of invalid objects in the epistemic uncertainty estimation. 

\noindent
\textbf{Undetected Number Prediction Network.} 
When estimating uncertainty for active object detection, previous works~\cite{wu2022entropy, yang2024plug, luo2023exploring} typically focus only on the uncertainty within the detected objects. However, the uncertainty of undetected objects is equally important, especially in the context of active learning for indoor 3D object detection, where model performance is often hindered by the scarcity of training samples and severe class imbalance. As shown by the mean average recall (mAR) scores in Tab.~\ref{tab: motivation of uncertainty calibration}, more than half of the ground truth objects remain undetected in both 3D indoor datasets. This observation highlights the need to account for the uncertainty of undetected objects. To address this, we design a lightweight network to estimate the number of undetected objects.

% \jy{
% Although some prior works~\cite{wu2022entropy} have studied uncertainty within detected objects, the uncertainty beyond detected proposals remains underexplored. 
% Shown by the mean average recall (mAR) scores in Table~\ref{tab: motivation of uncertainty calibration}, more than half ground truths remain undetected in both 3D indoor datasets during the early stages, primarily due to the scarcity of training samples and severe class imbalance.
% Hence, it is essential to exploit the undetected uncertainty, particularly for 3D indoor tasks.
% % it is essential to exploit this undetected uncertainty during sample selection, as complex indoor layouts and  pose additional challenges for the detection model.
% To capture epistemic uncertainty within undetected regions, we design a lightweight network to estimate the number of the undetected objects~\cite{nakamura2024active}. }

As depicted in Fig.~\ref{fig: architecture}, we first extract the feature map $\mathbf{f}_i$ 
from %of
the point cloud $\mathcal{P}_i$ using the feature extractor.
% , denoted as $\mathbf{f}_i$. 
We then apply global average pooling (GAP) on $\mathbf{f}_i$ to obtain the global feature $\mathbf{g}_i$. Simultaneously, we create a masked feature map by masking out the regions corresponding to the detected bounding boxes produced by RoI refinement. This masked feature map is passed through GAP to yield the undetected feature $\mathbf{u}_i$. We concatenate these two features, $\mathbf{g}_i$ and $\mathbf{u}_i$, and use the concatenation as input to our undetected object number prediction network $\Theta_{\text{undet}}$, a lightweight network consisting of three-layer MLPs. The output of the network, \ie $\Theta_{\text{undet}}([\mathbf{g}_i; \mathbf{u}_i])$ represents the uncertainty score for undetections, denoted as $\mathbf{U}_{\text{undet}}$. This output is compared with the true number of undetected objects, computed based on the detected objects $\mathfrak{B}_i$ and the ground-truth objects $\mathfrak{B}_i^*$, to compute the loss for training $\Theta_{\text{undet}}$:
\begin{equation}
    \label{equ: loss}
    \mathcal{L}_\text{undet} = \sum_{i=1}^{|\mathcal{D}_L|} (\Theta_{\text{undet}}([\mathbf{g}_i; \mathbf{u}_i])-\mathcal{N}(\mathfrak{B}_i, \mathfrak{B}_i^*))^2,
\end{equation}
where $\mathcal{N}(\cdot,\cdot)$ is a function that computes the number of undetected objects, with details provided in the supplementary material.
Additionally, to enhance the generalizability of the network $\Theta_{\text{undet}}$, we employ random masking of the predicted boxes during the training phase, encouraging the network to better capture the undetectability in point clouds.

% As depicted in Figure~\ref{fig: architecture}, we first extract global feature $\mathbf{g}_{\text{glo},i}$ by passing the feature map $\mathfrak{m}_i$ through global average pooling (GAP). Similarly, we retrieve the undetected feature $\mathbf{g}_{\text{und},i}$ from the masked feature map $\mathcal{M}(\mathfrak{m}_i)$ followed by GAP, where $\mathcal{M}(\cdot)$ masks 
% % the region inside the detected bounding boxes.  
% detected bounding boxes.
% The network then concatenates $\mathbf{g}_{\text{glo},i}$ and $\mathbf{g}_{\text{und},i}$ as inputs, with a loss function defined as:
% \begin{equation}
%     \label{equ: loss}
%     \mathcal{L}_{und} = (\bm{F}_{\text{und}}(\{\mathbf{g}_{\text{glo},i}; \mathbf{g}_{\text{und},i}\})-\mathcal{U}(\hat{\mathfrak{B}}_i, \mathfrak{B}_i))^2,
% \end{equation}
% where $\bm{F}_{\text{und}}(\cdot)$ is the proposed network, and $\mathcal{U}(\cdot,\cdot)$ computes the number of undetected objects. 
% The uncertainty score related to undetectability is directly defined as the outputs of the network, $\ie$, $\mathbf{U}_\text{und}(\mathcal{P}_i):=\bm{F}_\text{und}(\{\mathbf{g}_{\text{glo},i}; \mathbf{g}_{\text{und},i}\})$.
% Furthermore, to enhance the generalizablity of this lightweight network, we exploit random masking of the predicted boxes during the network training phase, encouraging it to better perceive the undetectability within the point clouds.

\vspace{0.02in}
\noindent
\textbf{Unified Uncertainty Score.} 
After obtaining the two uncertainty scores, $\mathbf{U}_{\text{det}}$ and $\mathbf{U}_{\text{undet}}$, we aim to combine them into a unified uncertainty score. However, these two scores are on different scales, making their direct product infeasible. To address this issue, we introduce a normalization step, inspired by \cite{nakamura2024active}, using a hyperparameter $k$. Specifically, we normalize each score as follows: 
\begin{equation}
    \label{equ: normalize}
    \tilde{\mathbf{U}}_{\star}(\mathcal{P}_i) = \text{max}(0,\frac{\mathbf{U}_{\star}(\mathcal{P}_i)-\mu(\mathbf{U}_{\star})}{k\cdot \sigma(\mathbf{U}_{\star})}+0.5),
\end{equation}
where ${\star} \in \{\text{det}, \text{undet}\}$, and $\mu(\cdot)$ and $\sigma(\cdot)$ denote the mean and standard deviation, respectively. Once the scores are normalized, we compute their product to yield the unified uncertainty score.

% The aforementioned scores, $\mathbf{U}_\text{loc}(\mathcal{P}_i)$ and $\mathbf{U}_\text{und}(\mathcal{P}_i)$, are then combined to form the final unified uncertainty score. However, their differing value ranges can impact the overall score quality. 
% To address this, inspired by~\cite{nakamura2024active}, we introduce a hyper-parameter $k$ to normalize each score:
% \begin{equation}
%     \label{equ: normalize}
%     \tilde{\mathbf{U}}_{\star}(\mathcal{P}_i) = \text{max}(0,\frac{\mathbf{U}_{\star}(\mathcal{P}_i)-\mu(\mathbf{U}_{\star})}{k\cdot \sigma(\mathbf{U}_{\star})}+0.5),
% \end{equation}
% where ${\star}\in \{\text{loc}, \text{und}\}$, and $\mu(\cdot)$ and $\sigma(\cdot)$ are mean and standard deviation. The unified uncertainty score is computed as the product of individual normalized metrics $\tilde{\mathbf{U}}_{\star}(\mathcal{P}_i)$. Finally, we select the top $B_r$ samples to constitute the candidate pool $\mathcal{D}_r^{\text{cand}}$ based on this unified uncertainty score.

%-------------------------------------------------------------------------
\subsection{Prototype-based Diversity Optimization}
\label{sec: prototype-based diversity optimization}

As illustrated in Fig.~\ref{fig: teaser}, indoor environments exhibit significant \textit{intra-class variances} and \textit{scene-type diversity}, both of which motivate our approach to the diversity criterion. In 3D indoor scenes, the intra-class variances can vary greatly across object categories due to the differences in shape and occlusion. For example, the variance in `chairs' is typically high due to the wide variety of shapes and designs, whereas `doors' tend to have lower variance due to their more standardized geometry. %because they follow more standardized forms. 
This variability makes methods that model each class with a fixed number of prototypes~\cite{xu2023generalized, zhao2021few} insufficient to capture the intra-class distributions. To address this, we propose a novel Class-aware Adaptive Prototype (CAP) bank, which dynamically allocates a varying number of prototypes for each class, inferred %learned
from the predicted object proposals. 
%underlying object embeddings.
%
Additionally, we observe that indoor scenes naturally form distinct clusters based on the distribution of object categories. For instance, bathrooms typically contain toilets and bathtubs, while bedrooms feature beds and dressers. 
Through %Leveraging
this observation, we represent each scene as a histogram of the prototypes corresponding to the predicted object categories within the scene. 
%To further capture scene-type diversity, we apply k-means++~\cite{wu2012advances} to organize these histograms into several scene-type clusters. 
As both intra-class variances and scene-type diversity can be modeled using the CAP bank, we frame diverse sample selection as an optimization problem that jointly maximizes both aspects, and propose an efficient solution.

\noindent
\textbf{Class-aware Adaptive Prototype Bank.} 
The design of our Class-aware Adaptive Prototype (CAP) bank is inspired by infinite mixture prototypes (IMP)~\cite{allen2019infinite}, which was originally proposed for few-shot learning. IMP allows for inferring the number of clusters from the data, rather than relying on a pre-defined number of clusters.
To build the CAP bank, we first iterate over the initial labeled set $D_L^{r-1}$ to initialize one prototype per class. Specifically, we leverage the ground-truth annotations to extract features for all objects belonging to the $c$-th class in $D_L^{r-1}$, and compute their average as to form the initial prototype $\bm{\upmu}_{1,c}$ for class $c$. 
%To build the CAP bank, we first iterate over the initial labeled set $D_L^0$ to initialize one prototype per class. Specifically, we leverage the ground-truth annotations to extract features for all objects belonging to the $c$-th class in $D_L^0$, and compute their average as to form the initial prototype $\bm{\upmu}_{1,c}$ for class $c$. 
%\zn{Subsequently, we iterate over the unlabeled dataset $\mathcal{D}_U$ for $R$ active learning rounds and update the prototypes for each class,}
Subsequently, we iterate over the unlabeled dataset $\mathcal{D}_U$ and update the prototypes for each class, 
resulting in our CAP bank, denoted as $\mathbf{M} = \{\{\bm{\upmu}_{k,c}\}_{k=1}^{M_c}\}_{c=1}^C$, where $\bm{\upmu}_{k,c}$ represents the $k$-th prototype for class $c$, $M_c$ is the number of prototypes for class $c$, which may vary between classes, and $C$ is the total number of classes. 
%\textcolor{red}{In the $r$-th active learning round, we use the detection model trained on $\mathcal{D}_L^{r-1}$ to generate object proposals and extract their features and predicted labels.} 
In each iteration (one batch), we use the detection model %trained on $\mathcal{D}_L^0$
from the previous round to generate object proposals and extract their features and predicted class labels.
For each predicted object, we compute the cosine similarity between its feature embedding $\mathbf{o}_j$ and each of the existing prototypes in its predicted class. This similarity score is formulated as:
\begin{equation}
    \label{equ: similiarity}
    \text{Sim}(\mathbf{o}_j, \bm{\upmu}_{k,c}) = \dfrac{\mathbf{o}_j^T\cdot \bm{\upmu}_{k,c}}{||\mathbf{o}_j||_2 ||\bm{\upmu}_{k,c}||_2} \in [-1,1].
\end{equation} 
After calculating the similarity scores, each object is either assigned to an existing prototype or used to form a new prototype, according to the following assignment rule:
\begin{equation}
    \label{equ: prototype assignment}
    \mathcal{A}(\mathbf{o}_j) =
      \begin{cases}
         M_c + 1, ~~\text{if $\underset{k}{\max}~\text{Sim}(\mathbf{o}_j, \bm{\upmu}_{k,c}) < \tau_{\text{sim}}$,}  \\
         \underset{k}{\arg \max}~\text{Sim}(\mathbf{o}_j, \bm{\upmu}_{k,c}), ~~~~~~~\text{otherwise}.
      \end{cases}
\end{equation}
Here, $\tau_{\text{sim}}$ is similarity threshold, which is set to 0.3 by default. 
Finally, the prototypes - whether assigned or newly formed - are updated using the new object proposals in each batch.
However, since many predicted bounding boxes might be invalid, especially in the early stages of training (\eg, as shown in Tab.~\ref{tab: motivation of uncertainty calibration}), we seek to mitigate the influence of these invalid proposals when updating the prototypes. To this end, we incorporate the IoU score from Eq.~\ref{equ: point cloud entropy} into the prototype update process, effectively weighing each object’s contribution to the prototype updates as follows:
\begin{equation}
    \label{equ: prototype updating scheme}
    \bm{\upmu}_{k,c} = \dfrac{\sum_{j=1}^{|\mathfrak{B}_i|}{\mathds{1}\{\mathcal{A}(\mathbf{o}_j), k\}}\cdot \text{IoU}_j \cdot \mathbf{o}_j}{\sum_{j=1}^{|\mathfrak{B}_i|}{\mathds{1}\{\mathcal{A}(\mathbf{o}_j), k\}}\cdot \text{IoU}_j},
\end{equation}
where the indicator function $\mathds{1}(m,n)$ equals to 1 if $m=n$, and 0 otherwise. 
The complete CAP bank creation procedure is outlined in Algo. \textcolor{green}{1} of the supplementary material.

\noindent
\textbf{Prototype-based Diversity Optimization.} 
% With the generated CAP bank, we can represent each class as a set of prototypes, and model objects as histograms of these prototypes.  
With the generated CAP bank, we can characterize each class through a set of prototypes, and further represent objects as histograms of these prototypes.
These histograms allow us to capture the intra-class variance for each class, as objects within the same class are reflected in the variation of their corresponding histograms. 
Furthermore, based on the observation that each scene can be characterized by the distribution of object categories, which distinguishes one scene type from another, we also model scenes as histograms of prototypes. Therefore, this histogram-based representation implicitly captures the scene type. 
Since both intra-class variance and scene-type diversity are encoded in the prototype histograms, we design a novel and efficient optimization method to quantify the diversity criterion by considering both these two aspects. Specifically, we formulate the optimization for diverse sample selection as follows:
% \begin{equation}
%     \label{equ: original optimization}
%     \begin{aligned}
%      \max_{\mathcal{D}_{r} \subset \mathcal{D}_r^{\text{cand}}} \quad & \underbrace{E(\mathbf{h}(\mathcal{D}_{r}))}_{\text{Variety}} + \beta \underbrace{||\mathbf{h}(\mathcal{D}_{r})||_1}_{\text{holism}},  \\
%     \textrm{s.t.} \quad  & \mathbf{h}(\mathcal{D}_{r}) = \underset{\mathcal{P}_i \in \mathcal{D}_r}{\sum} \mathbf{h}(\mathcal{P}_{i}) = \underset{\mathcal{P}_i \in \mathcal{D}_r}{\sum} \underset{\mathbf{o}_j \in \mathcal{P}_i}{\sum} \mathbf{h}(\mathbf{o}_{j}), \\%~~~~ |\mathcal{D}_r| = B_r, \\
%     & \mathbf{h}(\mathbf{o}_{j}) = \sum_{c=1}^C \sum_{k=1}^{M_c} [\text{one\_hot}(\mathcal{A}(\mathbf{o}_j))]^{c,k},
%        \\
%     \end{aligned}
% \end{equation}
\begin{equation}
\label{equ: original optimization}
    \begin{aligned}
     \max_{\mathcal{D}_{r} \subset \mathcal{D}_r^{\text{cand}}} \quad & E(\mathbf{h}(\mathcal{D}_{r})) + \beta ||\mathbf{h}(\mathcal{D}_{r})||_1,  \\
    \textrm{s.t.} \quad  & \mathbf{h}(\mathcal{D}_{r}) = \underset{\mathcal{P}_i \in \mathcal{D}_r}{\sum} \mathbf{h}(\mathcal{P}_{i}) = \underset{\mathcal{P}_i \in \mathcal{D}_r}{\sum} \underset{\mathbf{o}_j \in \mathcal{P}_i}{\sum} \mathbf{h}(\mathbf{o}_{j}), \\%~~~~ |\mathcal{D}_r| = B_r, \\
    & \mathbf{h}(\mathbf{o}_{j}) = \sum_{c=1}^C \sum_{k=1}^{M_c} [\text{one\_hot}(\mathcal{A}(\mathbf{o}_j))]^{c,k},
    \end{aligned}
    % \vspace{-0.05in}
\end{equation}
where $\mathbf{h}(\mathbf{o}_{j})$ represents the assignment histogram of $j$-th object \textit{w.r.t.} the CAP bank $\mathbf{M}$, while $\mathbf{h}(\mathcal{P}_{i})$ and $\mathbf{h}(\mathcal{D}_{r})$ represent the histograms of all objects in the individual point cloud $\mathcal{P}_i$ and the selected labeled samples $\mathcal{D}_r$ at the $r$-th round, respectively. 
%The two terms in this objective function are designed to select samples that are both \textbf{varied} and \textbf{holistic}.
The two terms in this objective function are designed to select samples that are both \textbf{diverse} and \textbf{holistic} across all prototypes in the CAP bank. 
The first term captures the intra-class variance for all classes, ensuring that the selected samples represent the diversity within each class as well as across different classes. The second term promotes scene-level diversity by encouraging coverage across different object categories, which implicitly increases the diversity in scene types.
Given that the first term (entropy) and the second term ($l_1$-norm) have different magnitudes, we introduce a hyperparameter $\beta$ to rescale the second term. In our experiments, we set $\beta = 0.01$.

However, directly solving the optimization in Eq.~\ref{equ: original optimization} is computationally expensive due to the combinatorial search over $|\mathcal{D}_{r}^{\text{cand}}|$ samples.
%\textcolor{red}{However, it is computationally intractable to solve this optimization directly, as it rquires the combinatorial search over $|\mathcal{D}_{r}^{\text{cand}}|$ samples.} 
%In order to effectively 
To efficiently narrow the search space while maintaining solution quality, we %highlight
leverage one key \textit{property} of the objective function over $S$ disjoint partitions $\{\mathcal{D}_{r,s}\}_{s=1}^S$ of $\mathcal{D}_r$, $\ie$, $\cup_s\mathcal{D}_{r,s}=\mathcal{D}_r$:
\vspace{-0.1in}
\begin{equation}
    \label{equ: property}
    \begin{aligned}
        & E(\mathbf{h}(\mathcal{D}_{r})) + \beta ||\mathbf{h}(\mathcal{D}_{r})||_1 \\
      \overset{c}{=} & \sum_{s=1}^S \{\frac{N_s}{N_{\text{total}}} E(\mathbf{h}(\mathcal{D}_{r,s})) + \beta ||\mathbf{h}(\mathcal{D}_{r,s})||_1 \} + \mathcal{O}(\frac{N_{\text{inter}}}{N_{\text{total}}}),
    \end{aligned}  
\end{equation}
% where $\mathbf{h}(\mathcal{D}_{r,s})$ represents the histograms of all objects in $s$-th partition $\mathcal{D}_{r,s}$, and $N_s$ and $N_{\text{total}}$ are the number of objects in the $s$-th partition $\mathcal{D}_{r,s}$ and selected labeled samples $\mathcal{D}_r$, and $N_{\text{inter}}$ is the number of objects that share the same prototypes but belong to different partitions.
where $\mathbf{h}(\mathcal{D}_{r,s})$ represents the histograms of all objects in $s$-th partition $\mathcal{D}_{r,s}$, $N_s$ and $N_{\text{total}}$ are the number of objects in the $s$-th partition $\mathcal{D}_{r,s}$ and selected labeled samples $\mathcal{D}_r$, $N_{\text{inter}}$ is the number of objects that share the same prototypes but belong to different partitions, and $\mathcal{O}(\cdot)$ is the Big O notation from mathematics.
%Moreover, due to the observation that indoor scenes form different clusters based on the distribution of the object categories contained, 
To align with the $S$ disjoint partitions of $\mathcal{D}_r$,
we use k-means++~\cite{wu2012advances} to group the candidate pool $\mathcal{D}_r^{\text{cand}}$ into $S$ clusters, denoted as $\{\mathcal{D}_{r,s}^{\text{cand}}\}_{s=1}^S$, each of which is approximately disjoint with each other \textit{w.r.t.} prototype assignments. 
% This guarantees that $N_{\text{inter}} = \smallO(N_{\text{total}})$, and then allows us to decompose the original optimization problem into $S$ sub-problems within each cluster $\mathcal{D}_{r,s}^{\text{cand}}$:
This guarantees that $\frac{N_{\text{inter}}}{N_{\text{total}}}\to 0$, and then allows us to decompose the original optimization problem into $S$ sub-problems within each cluster $\mathcal{D}_{r,s}^{\text{cand}}$:
\vspace{-0.1in}
\begin{equation}
    \begin{aligned} \label{equ: decomposed optimization}
     \max_{\mathcal{D}_{r,s} \subset \mathcal{D}_{r,s}^{\text{cand}}} \quad & E(\mathbf{h}(\mathcal{D}_{r,s})) + \beta~\dfrac{N_{\text{total}}}{{N_{s}}}~ ||\mathbf{h}(\mathcal{D}_{r,s})||_1,  \\
    \textrm{s.t.} \quad  & \mathbf{h}(\mathcal{D}_{r,s}) = \underset{\mathcal{P}_i \in \mathcal{D}_{r,s}}{\sum} \mathbf{h}(\mathcal{P}_{i}) = \underset{\mathcal{P}_i \in \mathcal{D}_{r,s}}{\sum} \underset{\mathbf{o}_j \in \mathcal{P}_i}{\sum} \mathbf{h}(\mathbf{o}_{j}), \\%~~~~ |\mathcal{D}_r| = B_r, \\
    & \mathbf{h}(\mathbf{o}_{j}) = \sum_{c=1}^C \sum_{k=1}^{M_c} [\text{one\_hot}(\mathcal{A}(\mathbf{o}_j))]^{c,k}.   \\
    %& |\mathcal{D}_{r,s}| = B_r/S.\\
    \end{aligned}
\end{equation}
We use greedy search to solve the $S$ sub-problems to obtain the diverse samples $\mathcal{D}_{r,s}$ for each cluster, and combine them together as the selected dataset $\mathcal{D}_r$ in the $r$-th active learning round. We provide the detailed derivation of Eq.~\ref{equ: property} in the supplementary material.

\section{Experiments}
\label{sec: experiments}

%-------------------------------------------------------------------------

\begin{table*}[t]
    \centering
    % \addtolength{\tabcolsep}{2pt}
    \vspace{-1ex}
    \addtolength{\tabcolsep}{5pt}
    \resizebox{.9\linewidth}{!}{% 
    \begin{tabular}{l l c cc cc}
        \toprule
         & &  & \multicolumn{2}{c}{SUN RGB-D} & \multicolumn{2}{c}{ScanNetV2}  \\
        \cmidrule(l){4-5}\cmidrule(l){6-7}
         & \multirow{-2}{*}{Methods}     & \multirow{-2}{*}{Presened at} & mAP@0.25 (\%) & mAP@0.50 (\%)  & mAP@0.25 (\%)  & mAP@0.50 (\%)  \\
        \midrule
        \parbox[t]{2mm}{\multirow{4}{*}{\rotatebox[origin=c]{90}{2D-Generic}}}  & \textbf{\textsc{Rand}} & - & 42.71 & 25.14 & 55.09 & 39.74 \\
         & \textbf{\textsc{Mc-Dropout}}~\cite{pmlr-v70-gal17a} & ICML'17 & 47.48 & 28.36 & 45.71 & 28.72 \\
         & \textbf{\textsc{Coreset}}~\cite{sener2018active}  & ICLR'18 & 46.99 & 28.12 & 55.34 & 40.08\\
         & \textbf{\textsc{Badge}}~\cite{ash2019deep} & ICLR'20 & 49.12 & 30.54 & 57.10 & 41.97 \\
        \midrule
        \parbox[t]{2mm}{\multirow{3}{*}{\rotatebox[origin=c]{90}{2D-OD}}} & 
        \textbf{\textsc{Cdal}}~\cite{agarwal2020contextual} & ECCV'20 & 50.86 & 30.75 & 57.99 & 42.33 \\
         & \textbf{\textsc{DivProto}}~\cite{wu2022entropy} & CVPR'22 & 52.31 & 32.54 & 59.01 & 43.10 \\
         & \textbf{\textsc{Ppal}}~\cite{yang2024plug} & CVPR'24 & 52.55 & 33.51 & 59.44 & 43.15 \\
        \midrule
         \parbox[t]{2mm}{\multirow{3}{*}{\rotatebox[origin=c]{90}{3D-OD}}} & \textbf{\textsc{Crb}}~\cite{luo2023exploring} & ICLR'23 & 50.89 & 31.30 & 58.73 & 42.71 \\
         & \textbf{\textsc{Kecor}}~\cite{luo2023kecor} & ICCV'23 & 51.07 & 31.41 & 58.54 & 42.56 \\
         & \textbf{\textsc{Ours}} & - & \textbf{56.11} & \textbf{35.42} & \textbf{61.77} & \textbf{45.93} \\
        \midrule
        \midrule
        & \textbf{\textsc{Oracle$^\dagger$}} & - & 64.31 & 47.43 & 72.65 & 58.81 \\
        \bottomrule
    \end{tabular}
    }
    %\begin{tablenotes}
        % \item {\small \hspace{-0.08in} $\dagger$: indicates the detection results using fully-annotated data, and serves as the upper bound for all active learning strategies.}
 %   \end{tablenotes}
    \vspace{-0.1in}
    \caption{\small{mAP@0.25 (\%) and mAP@0.50 (\%) scores of the proposed method and AL baselines on the SUN RGB-D and ScanNetV2 benchmarks with 10\% queried point clouds. CAGroup3D is used as the detection model for all approaches. $\dagger$: indicates the detection results using fully-annotated data, and serves as the upper bound for all active learning strategies.}}
\label{tab: experiment mAP}
\vspace{-2ex}
\end{table*}
  
% \begin{figure*}
%     \centering
%     \includegraphics[width=.95\linewidth]{experiment_plot.pdf}
%         \vspace{-0.1in}
%     \caption{\small{mAP@0.25 (\%) score of the proposed method and AL baselines on SUN RGB-D and ScanNetV2 benchmarks.}}
%     \label{fig: experiment}
% \end{figure*}
    %\vspace{-0.2in}

\subsection{Experimental Setups}
\textbf{3D Detection Datasets.} We evaluate our proposed method on two 3D indoor object detection benchmarks: SUN RGB-D~\cite{song2015sun} and ScanNet~\cite{dai2017scannet}. 
\textbf{SUN RGB-D} is an indoor dataset for 3D object detection, which contains 10,335 RGB-D images with 3D room layouts from four different sensors. Officially, they are split into 5,285 training samples and 5,050 testing samples. Following the standard evaluation protocol~\cite{qi2018frustum,qi2019deep,lahoud20172d}, we evaluate the performance on 10 most common categories.
\textbf{ScanNetV2} is a large 3D dataset with 1,513 RGB-D scans of 707 indoor scenes. These scans are split into 1,201 samples for training and 312 samples for testing, each of which is annotated with dense semantic segmentation masks. We follow~\cite{qi2019deep} to derive the axis-aligned bounding boxes without considering the orientation, and adopt 18 object categories for evaluation.

\noindent
\textbf{Implementation Details.} 
%For a fair comparison with prior hybrid works~\cite{wu2022entropy, yang2024plug, luo2023exploring}, we maintain the design~\cite{yang2024plug} that uses a budget expanding ratio $\delta > 1$ to first select the candidate pool of size $\delta \cdot B_r$ with Top-K uncertainty scores. Subsequently, diversity-based approach is applied to acquire the final $B_r$ active samples from the candidate pool.
We use the state-of-the-art CAGroup3D~\cite{wang2022cagroup3d} as the detection model, and implement our code base using OpenPCDet~\cite{openpcdet2020} toolkit. 
Following the training strategies in~\cite{wang2022cagroup3d}, we set the voxel size as $0.02$m.
For both datasets, the network is trained by AdamW optimizer~\cite{loshchilov2017decoupled} with an initial learning rate and weight decay rate of 0.001 and 0.0001 respectively. 
The gradnorm clip~\cite{zhang2020improved} is applied to stablize the training process.
For each active learning round, SUN RGB-D takes 48 epochs to train on the updated labeled dataset, and the learning rate decays on [32, 44] epochs with decay rate of 0.1.
ScanNet requires 120 epochs for training and decays the learning rate on [80, 110] epochs by the same decay rate.
All experiments are conducted using one 3090Ti NVIDIA GPU. We set the similarity threshold $\tau_{\text{sim}}$ to 0.3, balance ratio $\beta$ to 0.01, and budget expanding ratio $\delta$ to 6 empirically.

\noindent
\textbf{Evaluation.} 
Following the evaluation protocol in~\cite{qi2019deep, wang2022cagroup3d, misra2021end}, we report mean average precision (mAP) with different IoU thresholds, $\ie$, 0.25 and 0.50.
Additionally, to investigate the effectiveness of different active learning strategies, we record the mAP with 2\%, 4\%, 6\%, 8\%, 10\% of the training set for both datasets, where the first 2\% samples are selected randomly. In each active learning round, 2\% of the training point clouds are acquired from the rest unlabeled dataset via active learning strategies for annotation.

\begin{figure}
    \centering
    \includegraphics[width=\linewidth]{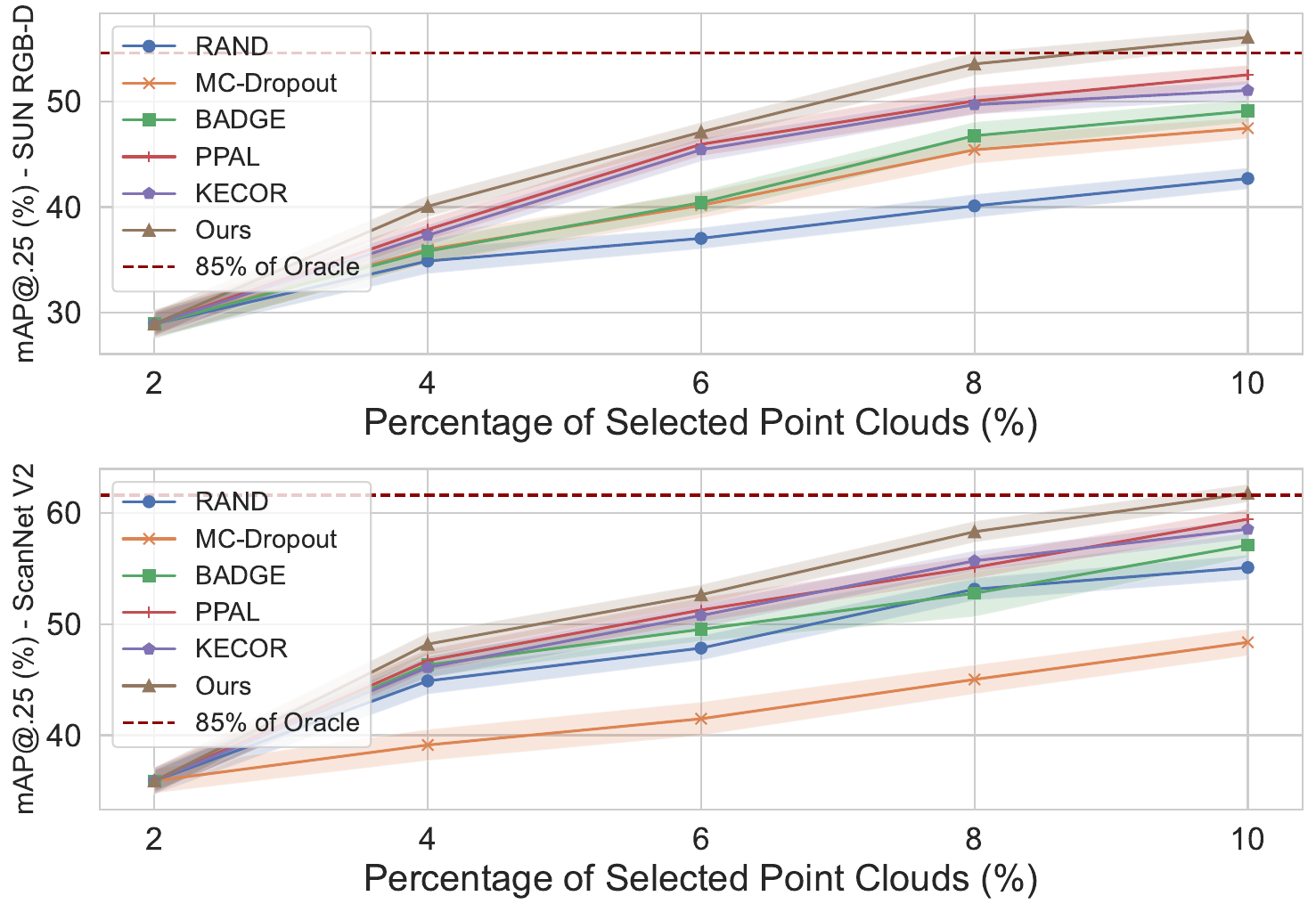}
        \vspace{-0.1in}
    \caption{\small{mAP@0.25 (\%) score of the proposed method and AL baselines on SUN RGB-D and ScanNetV2 benchmarks.}}
    \label{fig: experiment}
    \vspace{-0.15in}
\end{figure}

%-------------------------------------------------------------------------
%\subsection{Baselines}
\noindent
\textbf{Baselines.} 
We evaluate \textit{nine} AL baselines: random sampling (\textbf{\textsc{Rand}}), \textit{three} 2D generic AL strategies (\textbf{\textsc{Mc-Dropout}}, \textbf{\textsc{Coreset}}, \textbf{\textsc{Badge}}) representing uncertainty-based, diversity-based, and hybrid methods, \textit{three} 2D AL detection strategies (\textbf{\textsc{Cdal}}, \textbf{\textsc{DivProto}}, \textbf{\textsc{Ppal}}) adapted to 3D detection tasks by us, and \textit{two} state-of-the-art 3D outdoor AL detection strategies (\textbf{\textsc{Crb}} and \textbf{\textsc{Kecor}}).
\subsection{Experimental Results}
%To validate the effectiveness of our AL strategy, we compare with various baselines on two benchmark datasets.

\begin{figure*}
    \centering
    \includegraphics[width=\linewidth]{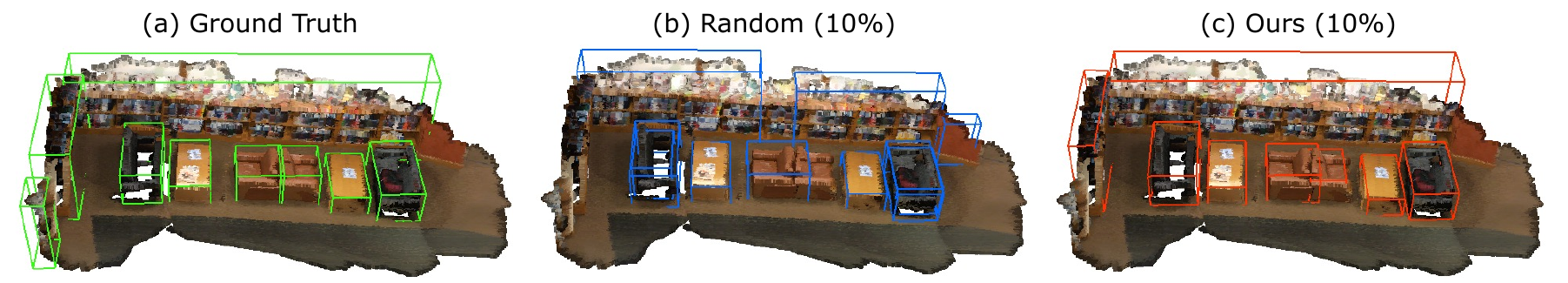}
        \vspace{-0.25in}
    \caption{\small{Qualitative comparison between random sampling and our proposed active learning method on ScanNetV2 val set.}}
    \label{fig: visualization}
    \vspace{-0.15in}
\end{figure*}

\noindent
\textbf{SUN RGB-D.} 
% 1. general results
In Fig.~\ref{fig: experiment}, we compare our proposed method with different baselines through an increasing percentage of selected point clouds~\cite{luo2023exploring, yang2024plug}. As demonstrated by the results, our proposed strategy consistently outperforms all competing baselines by significant margins for different percentages of selected point clouds. 
% 2. 85% oracle
Notably, with only 10\% annotated data, our approach achieves over \textbf{85}\% of the fully-annotated performance (red dashed line in Fig.~\ref{fig: experiment}). 
% 3. specific comparison with one baseline on one budget setting
Specifically, as detailed in Tab.~\ref{tab: experiment mAP}, it surpasses the state-of-the-art 2D active learning strategy, \textbf{\textsc{Ppal}}, by 3.56\% in mAP@0.25 and 1.91\% in mAP@0.50, with a 10\% annotation budget. 
% 4. observation, 2D OD > 3D outdoor OD
To elaborate further, it is observed that AL methods for 2D object detection generally outperform those for 3D outdoor counterpart. We hypothesize that current 3D outdoor approaches overlook the intra-class variances across different categories, leading to redundancy in selecting active samples. This redundancy may hinder the capability of the learned detection model. 
% 5. class-wise performance
% Lastly, Tab.~\ref{tab: experiment AP} provides detailed AP scores for several challenging classes with great intra-class diversity. The geometric variances increase the difficulty of accurate detection when training with limited data. Compared to different baselines, our approach shows substantial performance improvements across all challenging classes, $\eg$, `Desk' and `Bookshelf'.

\noindent
\textbf{ScanNetV2.} To study the effectiveness of our proposed approach on point clouds with complicated room layouts, we conduct experiments using the ScanNetV2 dataset, a more comprehensive benchmark for indoor environments. The results, illustrated in Fig.~\ref{fig: experiment} and Tab.~\ref{tab: experiment mAP}, demonstrate the performance comparison across varying annotation budgets. Our method surpasses all existing AL approaches by a huge margin, achieving approximately \textbf{85}\% of the fully-annotated performance with 10\% annotation budget.
% As shown in Tab.~\ref{tab: experiment AP}, previous approach struggle to consistently enhance detection quality across classes in complex indoor scenes. Specifically, mAP@0.25 decreases by 3.43\% and 1.69\% for \textbf{\textsc{Ppal}} and \textbf{\textsc{Kecor}}, respectively. In contrast, our approach effectively addresses this challenge through improved uncertainty estimation and more effective diversity formulation, leading to substantial performance gains across all classes.
% \torevise{[TODO: discuss Fig.4]}
In Fig.~\ref{fig: visualization}, we present a qualitative comparison with random sampling. Random sampling captures only partial object characteristics (\eg, `bookshelf') due to limited training data. In contrast, our approach successfully detects most of the target objects in challenging scenarios by effectively selecting informative unlabeled samples based on uncertainty and diversity criteria. Additional qualitative results can be found in the supplementary materials.
%\textcolor{red}{Furthermore, Fig.~\ref{fig: visualization} presents the qualitative results on ScanNetV2 using random sampling and the proposed strategy with 10\% annotation data. Random sampling only captures partial object characteristics (\eg, `bookshelf') due to limited training data. In contrast, our approach successfully detects most of the target objects in this challenging scenarios, by effectively estimating uncertainty and leveraging intra-class variances and scene-type diversity. We provide more visualization results in the supplementary materials.}

%-------------------------------------------------------------------------
\subsection{Ablation Study}
We ablate some critical designs
% of our method %on SUN RGB-D and ScanNetV2 
for an in-depth exploration.

\begin{table}[t]
    \begin{center}
    \resizebox{\linewidth}{!}{
    \begin{tabular}{c|c||cc||cc}
    \hline\hline
         \multirow{2}{*}{\textbf{Uncertainty}} & \multirow{2}{*}{\textbf{Diversity}} & \multicolumn{2}{c||}{SUN RGB-D} & \multicolumn{2}{c}{ScanNetV2} \\
        \cline{3-6}
         &   & mAP@.25 & mAP@.50 & mAP@.25 & mAP@.50  \\
        \hline \hline
        \ding{55} & \ding{55} & 42.7  & 25.1  & 55.1  & 39.7 \\
        \hline
        \hline
        \ding{55} & \ding{51} & 54.5 & 33.9  & 58.7 & 42.9 \\
        \hline
        w/ IoU score & \ding{51} & 55.6  & 34.7  & 60.8  & 43.8 \\
        \hline
        w/ $\mathbf{U}_{\text{undet}}$  & \ding{51} & 55.2  &  34.3 & 60.1  & 43.3 \\
        \hline
        \hline
        \ding{51} & \ding{55} & 54.3  & 33.5  &  58.4 & 42.8 \\
        \hline
        \ding{51} & w/ K-means & 55.3  &  34.1 &  60.4  & 43.5 \\
        \hline
        \ding{51} & w/ Greedy-O & 54.5  & 33.3  & 58.7  & 43.0 \\
        \hline
        \hline
        \ding{51} & \ding{51} & \textbf{56.1}  & \textbf{35.4}  & \textbf{61.8}  & \textbf{45.9} \\
    \hline \hline
    \end{tabular}}
    \end{center}
        \vspace{-0.25in}
    \caption{\small{Ablation studies of different designs in our proposed method on SUN RGB-D and ScanNetV2 datasets. }}
    \label{tab: ablation design}
    \vspace{-0.15in}
\end{table}

\noindent
%\textbf{Designs of proposed method.}
\textbf{Design Choices.}
We start by exploring the effectiveness of different designs in our proposed method, as shown in Tab.~\ref{tab: ablation design}. When equipped with the IoU score in Eq.~\ref{equ: point cloud entropy} and the uncertainty score for undetections $\mathbf{U}_{\text{undet}}$, 
%number of undetected objects in Eq.~\ref{equ: loss}, 
the mAP@0.25 increases by 1.1\% and 0.7\% on SUN RGB-D respectively, validating its effectiveness as an uncertainty-based strategy.
Additionally, we ablate the diversity branch through two alternatives: (1) Use K-means~\cite{wu2012advances} to select a fixed number of prototypes to replace CAP Bank, denoted as `\textit{K-means}'. (2) Apply greedy search to the original optimization problem as defined in Eq.~\ref{equ: original optimization}, denoted as `\textit{Greedy-O}'. 
It demonstrates that the mAP@0.50 of `\textit{K-means}' decreases by 2.4\% for ScanNetV2 dataset, which suggests that a fixed prototype number leads to a sub-optimal outcome. 
Furthermore, `\textit{Greedy-O}' solution only achieves minor performance improvements. A possible reason is that we restrict the searching space of the original optimization problem (Eq.~\ref{equ: original optimization}) to several regularized sub-spaces (Eq.~\ref{equ: decomposed optimization}), eliminating a vast number of sub-optimal solutions.
% When finding prototypes directly through `\textit{K-means}', %the mAP@0.25 score decreases by 0.8\% for SUN RGB-D dataset,
% the mAP@0.50 score decreases by 2.4\% for ScanNetV2 dataset, which suggests that a fixed number of prototypes leads to a sub-optimal outcome. Furthermore, `\textit{Greedy-O}' solution only achieves moderate performance improvements % with significant
% yet with larger fluctuations. A possible reason is that we restrict the searching space in the original optimization problem (Eq.~\ref{equ: original optimization}) to several regularized sub-spaces (Eq.~\ref{equ: decomposed optimization}) derived from %scene-type 
% disjoint clustering. This restriction eliminates a vast number of sub-optimal solutions, significantly enhancing the performance.

\begin{figure}
    \centering
    \includegraphics[width=\linewidth]{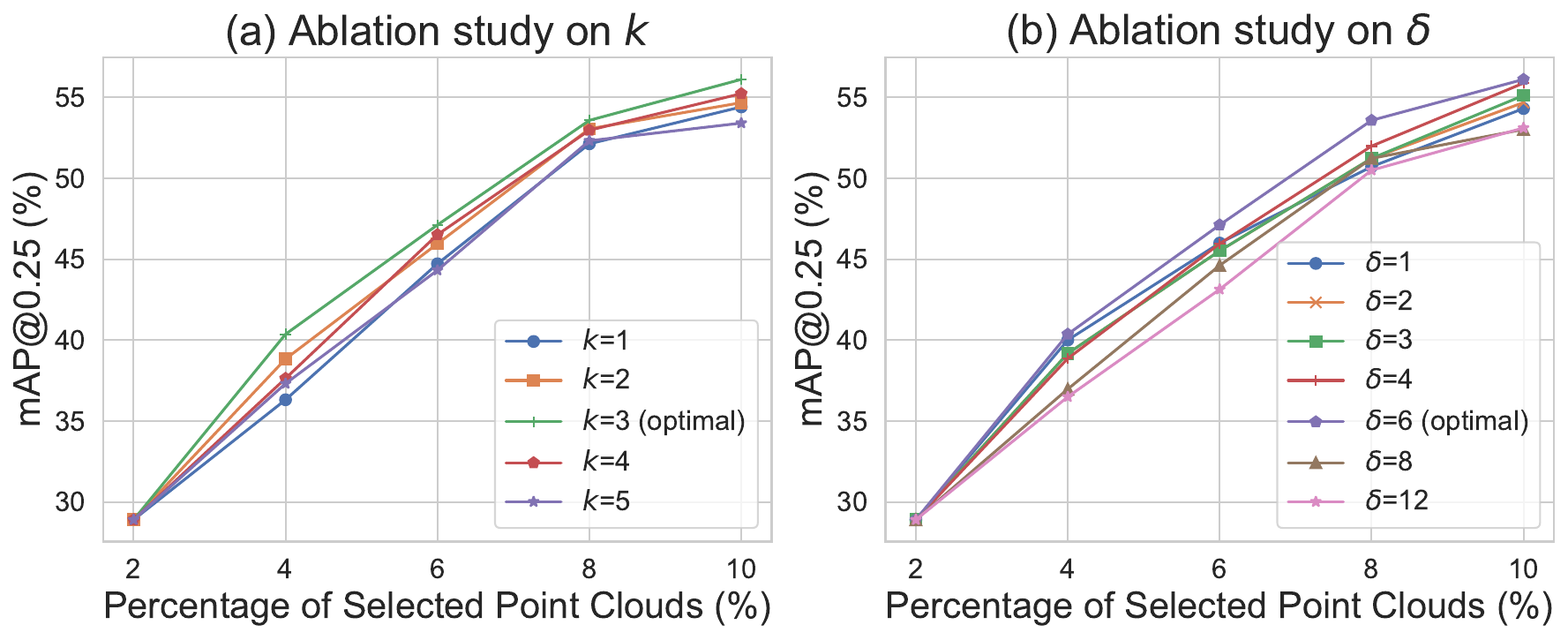}
        \vspace{-0.3in}
    \caption{\small{Effects of hyper-parameters in our proposed method on SUN RGB-D dataset.}}
    \label{fig: ablation hyperparameter}
    \vspace{-0.2in}
\end{figure}

\noindent
\textbf{Hyper-parameters.} 
In Fig.~\ref{fig: ablation hyperparameter}, we report the ablation study of two hyper-parameters in our proposed method, which are the scale $k$ in Eq.~\ref{equ: normalize} and the budget expanding ratio $\delta$ for determining the candidate pool~\cite{yang2024plug}. 
Overall, the proposed method is relatively insensitive to the choices of hyper-parameters. 
As depicted in Fig.~\ref{fig: ablation hyperparameter} (a), the best performance is reached with a moderate scale value of $k=3$, which effectively suppresses extreme values while preserving essential information from different sources of uncertainty. 
Additionally, we observe that excessively large values of $\delta$ will significantly degrade model performance in the early stages, which is shown in Fig.~\ref{fig: ablation hyperparameter} (b). 
% We hypothesize that for big $\delta$, there are numerous samples that the model is already certain about within the candidate pool, thereby harming the quality of final selected samples.
We hypothesize that for big $\delta$, the candidate pool contains many samples the model is already confident about, thereby harming the quality of the final selected samples.
Conversely, when $\delta$ is very small, the selection process is dominated by uncertainty, which restricts the impact of diversity branch.
We find that setting $\delta = 6$ achieves a better tradeoff between uncertainty- and diversity-based selection, optimizing overall performance.
%We find that the setting $\delta=6$ achieves a balance between uncertainty and diversity, optimizing the overall performance.

\begin{table}[t]
    \begin{center}
    \resizebox{\linewidth}{!}{
    \begin{tabular}{c|cc|cc}
    \hline
         \multirow{2}{*}{\textbf{Measures}} & \multicolumn{2}{c|}{SUN RGB-D} & \multicolumn{2}{c}{ScanNetV2} \\
        \cline{2-5}
           & mAP@.25 & mAP@.50 & mAP@.25 & mAP@.50  \\
           \hline
        Prob. Margin~\cite{roth2006margin}  & 55.8 & 35.0 & 61.5 & 45.5  \\
        \hline
         MC-Dropout~\cite{pmlr-v70-gal17a}  & 56.0 & 35.2 & 61.6 & 45.6  \\
           \hline
         Entropy~\cite{shannon1948mathematical}  & \textbf{56.1} & \textbf{35.4} & \textbf{61.8} & \textbf{45.9}  \\

    \hline
    \end{tabular}}
    \end{center}
        \vspace{-0.23in}
    \caption{\small{Ablation studies of different uncertainty measures in our proposed method on SUN RGB-D and ScanNetV2 datasets.}}
    \label{tab: ablation measurement}
    \vspace{-0.2in}
\end{table}

\noindent
%\textbf{Uncertainty Measurements.} 
\textbf{Uncertainty Measures.} 
We replace the Shannon entropy in Eq.~\ref{equ: point cloud entropy} with two alternative uncertainty measures: MC-Dropout~\cite{pmlr-v70-gal17a} and probability margin~\cite{roth2006margin}, and compare their performance in Tab.~\ref{tab: ablation measurement}. The results show that all three measures yield similar performance, with entropy performing slightly better. Therefore, we select entropy as our uncertainty measure, as it is both more efficient and effective.
%We highlight that the Shannon entropy $E(\mathbf{p}_j)$ in Eq.~\ref{equ: point cloud entropy} for the $j$-th object can be replaced with any uncertainty measure. Accordingly, we present the ablation studies of different uncertainty measures in Tab.~\ref{tab: ablation measurement}, including MC-Dropout~\cite{pmlr-v70-gal17a}, Shannon entropy~\cite{shannon1948mathematical}, and probability margin~\cite{roth2006margin}. We evaluate the performance on SUN RGB-D and ScanNet datasets in terms of mAP@0.25 and mAP@0.50. The results indicate that all three measures yield similar performance, though probability margin performs slightly lower than the other two. MC-Dropout is more stable but incurs higher memory costs due to  multiple forward passes. Therefore, we select entropy as our uncertainty measure, achieving a trade-off between effectiveness and computational efficiency.

%-------------------------------------------------------------------------
%\subsection{Qualitative Results and Analysis}
\section{Conclusion}
\label{sec: conclusion}
This paper presents a novel active learning approach for 3D indoor object detection, improving annotation efficiency by combining uncertainty estimation with diversity-driven sampling. Specifically, our approach estimates epistemic uncertainty, accounting for both inaccurate detections and undetected objects, and selects diverse samples through an optimization method based on a Class-aware Adaptive Prototype bank. Our method significantly outperforms state-of-the-art baselines, achieving over 85\% of fully-supervised performance with just 10\% of the annotation budget.

\section{Acknowledgment}
This research is supported by the Ministry of Education, Singapore, under its MOE Academic Research Fund Tier 1 -- SMU-SUTD Internal Research Grant (SMU-SUTD 2023\_02\_09), it is also supported by the Agency for Science, Technology and Research (A*STAR) under its MTC Programmatic Funds (Grant No. M23L7b0021).

{
    \small
    \bibliographystyle{ieeenat_fullname}
    \bibliography{main}

\begin{thebibliography}{60}
\providecommand{\natexlab}[1]{#1}
\providecommand{\url}[1]{\texttt{#1}}
\expandafter\ifx\csname urlstyle\endcsname\relax
  \providecommand{\doi}[1]{doi: #1}\else
  \providecommand{\doi}{doi: \begingroup \urlstyle{rm}\Url}\fi

\bibitem[Agarwal et~al.(2020)Agarwal, Arora, Anand, and Arora]{agarwal2020contextual}
Sharat Agarwal, Himanshu Arora, Saket Anand, and Chetan Arora.
\newblock Contextual diversity for active learning.
\newblock In \emph{Computer Vision--ECCV 2020: 16th European Conference, Glasgow, UK, August 23--28, 2020, Proceedings, Part XVI 16}, pages 137--153. Springer, 2020.

\bibitem[Allen et~al.(2019)Allen, Shelhamer, Shin, and Tenenbaum]{allen2019infinite}
Kelsey Allen, Evan Shelhamer, Hanul Shin, and Joshua Tenenbaum.
\newblock Infinite mixture prototypes for few-shot learning.
\newblock In \emph{International conference on machine learning}, pages 232--241. PMLR, 2019.

\bibitem[Arnold et~al.(2019)Arnold, Al-Jarrah, Dianati, Fallah, Oxtoby, and Mouzakitis]{arnold2019survey}
Eduardo Arnold, Omar~Y Al-Jarrah, Mehrdad Dianati, Saber Fallah, David Oxtoby, and Alex Mouzakitis.
\newblock A survey on 3d object detection methods for autonomous driving applications.
\newblock \emph{IEEE Transactions on Intelligent Transportation Systems}, 20\penalty0 (10):\penalty0 3782--3795, 2019.

\bibitem[Ash et~al.(2019)Ash, Zhang, Krishnamurthy, Langford, and Agarwal]{ash2019deep}
Jordan~T Ash, Chicheng Zhang, Akshay Krishnamurthy, John Langford, and Alekh Agarwal.
\newblock Deep batch active learning by diverse, uncertain gradient lower bounds.
\newblock \emph{arXiv preprint arXiv:1906.03671}, 2019.

\bibitem[Citovsky et~al.(2021)Citovsky, DeSalvo, Gentile, Karydas, Rajagopalan, Rostamizadeh, and Kumar]{citovsky2021batch}
Gui Citovsky, Giulia DeSalvo, Claudio Gentile, Lazaros Karydas, Anand Rajagopalan, Afshin Rostamizadeh, and Sanjiv Kumar.
\newblock Batch active learning at scale.
\newblock \emph{Advances in Neural Information Processing Systems}, 34:\penalty0 11933--11944, 2021.

\bibitem[Dai et~al.(2017)Dai, Chang, Savva, Halber, Funkhouser, and Nie{\ss}ner]{dai2017scannet}
Angela Dai, Angel~X. Chang, Manolis Savva, Maciej Halber, Thomas Funkhouser, and Matthias Nie{\ss}ner.
\newblock Scannet: Richly-annotated 3d reconstructions of indoor scenes.
\newblock In \emph{Proc. Computer Vision and Pattern Recognition (CVPR), IEEE}, 2017.

\bibitem[Elhamifar et~al.(2013)Elhamifar, Sapiro, Yang, and Sasrty]{elhamifar2013convex}
Ehsan Elhamifar, Guillermo Sapiro, Allen Yang, and S~Shankar Sasrty.
\newblock A convex optimization framework for active learning.
\newblock In \emph{Proceedings of the IEEE International Conference on Computer Vision}, pages 209--216, 2013.

\bibitem[Feng et~al.(2019)Feng, Wei, Rosenbaum, Maki, and Dietmayer]{feng2019deep}
Di Feng, Xiao Wei, Lars Rosenbaum, Atsuto Maki, and Klaus Dietmayer.
\newblock Deep active learning for efficient training of a lidar 3d object detector.
\newblock In \emph{2019 IEEE Intelligent Vehicles Symposium (IV)}, pages 667--674. IEEE, 2019.

\bibitem[Gal et~al.(2017)Gal, Islam, and Ghahramani]{pmlr-v70-gal17a}
Yarin Gal, Riashat Islam, and Zoubin Ghahramani.
\newblock Deep {B}ayesian active learning with image data.
\newblock In \emph{Proceedings of the 34th International Conference on Machine Learning}, pages 1183--1192. PMLR, 2017.

\bibitem[Gashler and Martinez(2011)]{gashler2011temporal}
Mike Gashler and Tony Martinez.
\newblock Temporal nonlinear dimensionality reduction.
\newblock In \emph{The 2011 International Joint Conference on Neural Networks}, pages 1959--1966. IEEE, 2011.

\bibitem[Geiger et~al.(2013)Geiger, Lenz, Stiller, and Urtasun]{geiger2013vision}
Andreas Geiger, Philip Lenz, Christoph Stiller, and Raquel Urtasun.
\newblock Vision meets robotics: The kitti dataset.
\newblock \emph{The International Journal of Robotics Research}, 32\penalty0 (11):\penalty0 1231--1237, 2013.

\bibitem[Ghita et~al.(2024)Ghita, Antoniussen, Zimmer, Greer, Cre{\ss}, M{\o}gelmose, Trivedi, and Knoll]{ghita2024activeanno3d}
Ahmed Ghita, Bj{\o}rk Antoniussen, Walter Zimmer, Ross Greer, Christian Cre{\ss}, Andreas M{\o}gelmose, Mohan~M Trivedi, and Alois~C Knoll.
\newblock Activeanno3d--an active learning framework for multi-modal 3d object detection.
\newblock \emph{arXiv preprint arXiv:2402.03235}, 2024.

\bibitem[Gwak et~al.(2020)Gwak, Choy, and Savarese]{gwak2020generative}
JunYoung Gwak, Christopher Choy, and Silvio Savarese.
\newblock Generative sparse detection networks for 3d single-shot object detection.
\newblock In \emph{Computer Vision--ECCV 2020: 16th European Conference, Glasgow, UK, August 23--28, 2020, Proceedings, Part IV 16}, pages 297--313. Springer, 2020.

\bibitem[Han et~al.(2024)Han, Zhao, Chen, Ma, and Zhang]{han2024dual}
Yucheng Han, Na Zhao, Weiling Chen, Keng~Teck Ma, and Hanwang Zhang.
\newblock Dual-perspective knowledge enrichment for semi-supervised 3d object detection.
\newblock In \emph{Proceedings of the AAAI Conference on Artificial Intelligence}, pages 2049--2057, 2024.

\bibitem[Joshi et~al.(2009)Joshi, Porikli, and Papanikolopoulos]{joshi2009multi}
Ajay~J Joshi, Fatih Porikli, and Nikolaos Papanikolopoulos.
\newblock Multi-class active learning for image classification.
\newblock In \emph{2009 ieee conference on computer vision and pattern recognition}, pages 2372--2379. IEEE, 2009.

\bibitem[Kim et~al.(2021)Kim, Song, Jang, and Moon]{kim2021lada}
Yoon-Yeong Kim, Kyungwoo Song, JoonHo Jang, and Il-Chul Moon.
\newblock Lada: Look-ahead data acquisition via augmentation for deep active learning.
\newblock \emph{Advances in Neural Information Processing Systems}, 34:\penalty0 22919--22930, 2021.

\bibitem[Kye et~al.(2023)Kye, Choi, Byun, and Chang]{kye2023tidal}
Seong~Min Kye, Kwanghee Choi, Hyeongmin Byun, and Buru Chang.
\newblock Tidal: Learning training dynamics for active learning.
\newblock In \emph{Proceedings of the IEEE/CVF International Conference on Computer Vision}, pages 22335--22345, 2023.

\bibitem[Lahoud and Ghanem(2017)]{lahoud20172d}
Jean Lahoud and Bernard Ghanem.
\newblock 2d-driven 3d object detection in rgb-d images.
\newblock In \emph{Proceedings of the IEEE international conference on computer vision}, pages 4622--4630, 2017.

\bibitem[Li et~al.(2022)Li, Wang, Li, Xie, Sima, Lu, Qiao, and Dai]{li2022bevformer}
Zhiqi Li, Wenhai Wang, Hongyang Li, Enze Xie, Chonghao Sima, Tong Lu, Yu Qiao, and Jifeng Dai.
\newblock Bevformer: Learning bird’s-eye-view representation from multi-camera images via spatiotemporal transformers.
\newblock In \emph{European conference on computer vision}, pages 1--18. Springer, 2022.

\bibitem[Lin et~al.(2024)Lin, Liang, Deng, Cai, Jiang, Li, Jia, and Xu]{lin2024exploring}
Jinpeng Lin, Zhihao Liang, Shengheng Deng, Lile Cai, Tao Jiang, Tianrui Li, Kui Jia, and Xun Xu.
\newblock Exploring diversity-based active learning for 3d object detection in autonomous driving.
\newblock \emph{IEEE Transactions on Intelligent Transportation Systems}, 2024.

\bibitem[Liu et~al.(2021)Liu, Ding, Zhong, Li, Dai, and He]{liu2021influence}
Zhuoming Liu, Hao Ding, Huaping Zhong, Weijia Li, Jifeng Dai, and Conghui He.
\newblock Influence selection for active learning.
\newblock In \emph{Proceedings of the IEEE/CVF international conference on computer vision}, pages 9274--9283, 2021.

\bibitem[Loshchilov(2017)]{loshchilov2017decoupled}
I Loshchilov.
\newblock Decoupled weight decay regularization.
\newblock \emph{arXiv preprint arXiv:1711.05101}, 2017.

\bibitem[Luo et~al.(2023{\natexlab{a}})Luo, Chen, Fang, Zhang, Baktashmotlagh, and Huang]{luo2023kecor}
Yadan Luo, Zhuoxiao Chen, Zhen Fang, Zheng Zhang, Mahsa Baktashmotlagh, and Zi Huang.
\newblock Kecor: Kernel coding rate maximization for active 3d object detection.
\newblock In \emph{Proceedings of the IEEE/CVF International Conference on Computer Vision}, pages 18279--18290, 2023{\natexlab{a}}.

\bibitem[Luo et~al.(2023{\natexlab{b}})Luo, Chen, Wang, Yu, Huang, and Baktashmotlagh]{luo2023exploring}
Yadan Luo, Zhuoxiao Chen, Zijian Wang, Xin Yu, Zi Huang, and Mahsa Baktashmotlagh.
\newblock Exploring active 3d object detection from a generalization perspective.
\newblock \emph{arXiv preprint arXiv:2301.09249}, 2023{\natexlab{b}}.

\bibitem[Ma et~al.(2007)Ma, Derksen, Hong, and Wright]{ma2007segmentation}
Yi Ma, Harm Derksen, Wei Hong, and John Wright.
\newblock Segmentation of multivariate mixed data via lossy data coding and compression.
\newblock \emph{IEEE transactions on pattern analysis and machine intelligence}, 29\penalty0 (9):\penalty0 1546--1562, 2007.

\bibitem[Maisano et~al.(2018)Maisano, Tomaselli, Capra, Longo, and Puliafito]{maisano2018reducing}
Roberta Maisano, Valeria Tomaselli, Alessandro Capra, Francesco Longo, and Antonio Puliafito.
\newblock Reducing complexity of 3d indoor object detection.
\newblock In \emph{2018 IEEE 4th International Forum on Research and Technology for Society and Industry (RTSI)}, pages 1--6. IEEE, 2018.

\bibitem[Maturana and Scherer(2015)]{maturana2015voxnet}
Daniel Maturana and Sebastian Scherer.
\newblock Voxnet: A 3d convolutional neural network for real-time object recognition.
\newblock In \emph{2015 IEEE/RSJ international conference on intelligent robots and systems (IROS)}, pages 922--928. IEEE, 2015.

\bibitem[Meyer and Kuschk(2019)]{meyer2019automotive}
Michael Meyer and Georg Kuschk.
\newblock Automotive radar dataset for deep learning based 3d object detection.
\newblock In \emph{2019 16th european radar conference (EuRAD)}, pages 129--132. IEEE, 2019.

\bibitem[Mi et~al.(2022)Mi, Lin, Zhou, Shen, Luo, Sun, Cao, Fu, Xu, and Ji]{mi2022active}
Peng Mi, Jianghang Lin, Yiyi Zhou, Yunhang Shen, Gen Luo, Xiaoshuai Sun, Liujuan Cao, Rongrong Fu, Qiang Xu, and Rongrong Ji.
\newblock Active teacher for semi-supervised object detection.
\newblock In \emph{Proceedings of the IEEE/CVF conference on computer vision and pattern recognition}, pages 14482--14491, 2022.

\bibitem[Misra et~al.(2021)Misra, Girdhar, and Joulin]{misra2021end}
Ishan Misra, Rohit Girdhar, and Armand Joulin.
\newblock An end-to-end transformer model for 3d object detection.
\newblock In \emph{Proceedings of the IEEE/CVF international conference on computer vision}, pages 2906--2917, 2021.

\bibitem[Moses et~al.(2022)Moses, Jakkampudi, Danner, and Biega]{moses2022localization}
Aimee Moses, Srikanth Jakkampudi, Cheryl Danner, and Derek Biega.
\newblock Localization-based active learning (local) for object detection in 3d point clouds.
\newblock In \emph{Geospatial Informatics XII}, pages 44--58. SPIE, 2022.

\bibitem[Nakamura et~al.(2024)Nakamura, Ishii, and Yamashita]{nakamura2024active}
Yuzuru Nakamura, Yasunori Ishii, and Takayoshi Yamashita.
\newblock Active domain adaptation with false negative prediction for object detection.
\newblock In \emph{Proceedings of the IEEE/CVF Conference on Computer Vision and Pattern Recognition}, pages 28782--28792, 2024.

\bibitem[Nguyen and Smeulders(2004)]{nguyen2004active}
Hieu~T Nguyen and Arnold Smeulders.
\newblock Active learning using pre-clustering.
\newblock In \emph{Proceedings of the twenty-first international conference on Machine learning}, page~79, 2004.

\bibitem[Qi et~al.(2018)Qi, Liu, Wu, Su, and Guibas]{qi2018frustum}
Charles~R Qi, Wei Liu, Chenxia Wu, Hao Su, and Leonidas~J Guibas.
\newblock Frustum pointnets for 3d object detection from rgb-d data.
\newblock In \emph{Proceedings of the IEEE conference on computer vision and pattern recognition}, pages 918--927, 2018.

\bibitem[Qi et~al.(2019)Qi, Litany, He, and Guibas]{qi2019deep}
Charles~R Qi, Or Litany, Kaiming He, and Leonidas~J Guibas.
\newblock Deep hough voting for 3d object detection in point clouds.
\newblock In \emph{proceedings of the IEEE/CVF International Conference on Computer Vision}, pages 9277--9286, 2019.

\bibitem[Roth and Small(2006)]{roth2006margin}
Dan Roth and Kevin Small.
\newblock Margin-based active learning for structured output spaces.
\newblock In \emph{Machine Learning: ECML 2006: 17th European Conference on Machine Learning Berlin, Germany, September 18-22, 2006 Proceedings 17}, pages 413--424. Springer, 2006.

\bibitem[Sener and Savarese(2018)]{sener2018active}
Ozan Sener and Silvio Savarese.
\newblock Active learning for convolutional neural networks: A core-set approach.
\newblock In \emph{International Conference on Learning Representations}, 2018.

\bibitem[Shannon(1948)]{shannon1948mathematical}
Claude~Elwood Shannon.
\newblock A mathematical theory of communication.
\newblock \emph{The Bell system technical journal}, 27\penalty0 (3):\penalty0 379--423, 1948.

\bibitem[Shen and Stamos(2020)]{shen2020frustum}
Xiaoke Shen and Ioannis Stamos.
\newblock Frustum voxnet for 3d object detection from rgb-d or depth images.
\newblock In \emph{Proceedings of the IEEE/CVF winter conference on applications of computer vision}, pages 1698--1706, 2020.

\bibitem[Shen et~al.(2023)Shen, Geng, Yuan, Lin, Liu, Wang, Hu, Zheng, and Guo]{shen2023v}
Yichao Shen, Zigang Geng, Yuhui Yuan, Yutong Lin, Ze Liu, Chunyu Wang, Han Hu, Nanning Zheng, and Baining Guo.
\newblock V-detr: Detr with vertex relative position encoding for 3d object detection.
\newblock \emph{arXiv preprint arXiv:2308.04409}, 2023.

\bibitem[Sinha et~al.(2019)Sinha, Ebrahimi, and Darrell]{sinha2019variational}
Samarth Sinha, Sayna Ebrahimi, and Trevor Darrell.
\newblock Variational adversarial active learning.
\newblock In \emph{Proceedings of the IEEE/CVF international conference on computer vision}, pages 5972--5981, 2019.

\bibitem[Song et~al.(2015)Song, Lichtenberg, and Xiao]{song2015sun}
Shuran Song, Samuel~P Lichtenberg, and Jianxiong Xiao.
\newblock Sun rgb-d: A rgb-d scene understanding benchmark suite.
\newblock In \emph{Proceedings of the IEEE conference on computer vision and pattern recognition}, pages 567--576, 2015.

\bibitem[Team(2020)]{openpcdet2020}
OpenPCDet~Development Team.
\newblock Openpcdet: An open-source toolbox for 3d object detection from point clouds.
\newblock \url{https://github.com/open-mmlab/OpenPCDet}, 2020.

\bibitem[Wang et~al.(2022)Wang, Ding, Dong, Shi, Li, Li, Li, and Wang]{wang2022cagroup3d}
Haiyang Wang, Lihe Ding, Shaocong Dong, Shaoshuai Shi, Aoxue Li, Jianan Li, Zhenguo Li, and Liwei Wang.
\newblock Cagroup3d: Class-aware grouping for 3d object detection on point clouds.
\newblock \emph{Advances in Neural Information Processing Systems}, 35:\penalty0 29975--29988, 2022.

\bibitem[Wang et~al.(2024)Wang, Cheng, Zhao, Cheng, and Yang]{wang2024fly}
Jiangyi Wang, Zhongyao Cheng, Na Zhao, Jun Cheng, and Xulei Yang.
\newblock On-the-fly point feature representation for point clouds analysis.
\newblock \emph{arXiv preprint arXiv:2407.21335}, 2024.

\bibitem[Wang et~al.(2016)Wang, Zhang, Li, Zhang, and Lin]{wang2016cost}
Keze Wang, Dongyu Zhang, Ya Li, Ruimao Zhang, and Liang Lin.
\newblock Cost-effective active learning for deep image classification.
\newblock \emph{IEEE Transactions on Circuits and Systems for Video Technology}, 27\penalty0 (12):\penalty0 2591--2600, 2016.

\bibitem[Watson et~al.(2023)Watson, Sayed, Qureshi, Brostow, Vicente, Mac~Aodha, and Firman]{Watson_2023_CVPR}
Jamie Watson, Mohamed Sayed, Zawar Qureshi, Gabriel~J. Brostow, Sara Vicente, Oisin Mac~Aodha, and Michael Firman.
\newblock Virtual occlusions through implicit depth.
\newblock In \emph{Proceedings of the IEEE/CVF Conference on Computer Vision and Pattern Recognition (CVPR)}, pages 9053--9064, 2023.

\bibitem[Wu(2012)]{wu2012advances}
Junjie Wu.
\newblock \emph{Advances in K-means clustering: a data mining thinking}.
\newblock Springer Science \& Business Media, 2012.

\bibitem[Wu et~al.(2022)Wu, Chen, and Huang]{wu2022entropy}
Jiaxi Wu, Jiaxin Chen, and Di Huang.
\newblock Entropy-based active learning for object detection with progressive diversity constraint.
\newblock In \emph{Proceedings of the IEEE/CVF Conference on Computer Vision and Pattern Recognition}, pages 9397--9406, 2022.

\bibitem[Wu et~al.(2021)Wu, Liu, Huang, Lee, Su, Huang, and Hsu]{wu2021redal}
Tsung-Han Wu, Yueh-Cheng Liu, Yu-Kai Huang, Hsin-Ying Lee, Hung-Ting Su, Ping-Chia Huang, and Winston~H Hsu.
\newblock Redal: Region-based and diversity-aware active learning for point cloud semantic segmentation.
\newblock In \emph{Proceedings of the IEEE/CVF international conference on computer vision}, pages 15510--15519, 2021.

\bibitem[Xie et~al.(2020)Xie, Lai, Wu, Wang, Zhang, Xu, and Wang]{xie2020mlcvnet}
Qian Xie, Yu-Kun Lai, Jing Wu, Zhoutao Wang, Yiming Zhang, Kai Xu, and Jun Wang.
\newblock Mlcvnet: Multi-level context votenet for 3d object detection.
\newblock In \emph{Proceedings of the IEEE/CVF conference on computer vision and pattern recognition}, pages 10447--10456, 2020.

\bibitem[Xie et~al.(2021)Xie, Lai, Wu, Wang, Lu, Wei, and Wang]{xie2021venet}
Qian Xie, Yu-Kun Lai, Jing Wu, Zhoutao Wang, Dening Lu, Mingqiang Wei, and Jun Wang.
\newblock Venet: Voting enhancement network for 3d object detection.
\newblock In \emph{Proceedings of the IEEE/CVF International Conference on Computer Vision}, pages 3712--3721, 2021.

\bibitem[Xu et~al.(2023)Xu, Hu, Zhao, and Lee]{xu2023generalized}
Yating Xu, Conghui Hu, Na Zhao, and Gim~Hee Lee.
\newblock Generalized few-shot point cloud segmentation via geometric words.
\newblock In \emph{Proceedings of the IEEE/CVF International Conference on Computer Vision}, pages 21506--21515, 2023.

\bibitem[Yang et~al.(2023)Yang, Chen, Tian, Tao, Zhu, Zhang, Huang, Li, Qiao, Lu, et~al.]{yang2023bevformer}
Chenyu Yang, Yuntao Chen, Hao Tian, Chenxin Tao, Xizhou Zhu, Zhaoxiang Zhang, Gao Huang, Hongyang Li, Yu Qiao, Lewei Lu, et~al.
\newblock Bevformer v2: Adapting modern image backbones to bird's-eye-view recognition via perspective supervision.
\newblock In \emph{Proceedings of the IEEE/CVF Conference on Computer Vision and Pattern Recognition}, pages 17830--17839, 2023.

\bibitem[Yang et~al.(2024)Yang, Huang, and Crowley]{yang2024plug}
Chenhongyi Yang, Lichao Huang, and Elliot~J Crowley.
\newblock Plug and play active learning for object detection.
\newblock In \emph{Proceedings of the IEEE/CVF Conference on Computer Vision and Pattern Recognition}, pages 17784--17793, 2024.

\bibitem[Yang et~al.(2015)Yang, Ma, Nie, Chang, and Hauptmann]{yang2015multi}
Yi Yang, Zhigang Ma, Feiping Nie, Xiaojun Chang, and Alexander~G Hauptmann.
\newblock Multi-class active learning by uncertainty sampling with diversity maximization.
\newblock \emph{International Journal of Computer Vision}, 113:\penalty0 113--127, 2015.

\bibitem[Yuan et~al.(2021)Yuan, Wan, Fu, Liu, Xu, Ji, and Ye]{yuan2021multiple}
Tianning Yuan, Fang Wan, Mengying Fu, Jianzhuang Liu, Songcen Xu, Xiangyang Ji, and Qixiang Ye.
\newblock Multiple instance active learning for object detection.
\newblock In \emph{Proceedings of the IEEE/CVF Conference on Computer Vision and Pattern Recognition}, pages 5330--5339, 2021.

\bibitem[Zhang et~al.(2020)Zhang, Jin, Fang, and Wang]{zhang2020improved}
Bohang Zhang, Jikai Jin, Cong Fang, and Liwei Wang.
\newblock Improved analysis of clipping algorithms for non-convex optimization.
\newblock \emph{Advances in Neural Information Processing Systems}, 33:\penalty0 15511--15521, 2020.

\bibitem[Zhao et~al.(2021)Zhao, Chua, and Lee]{zhao2021few}
Na Zhao, Tat-Seng Chua, and Gim~Hee Lee.
\newblock Few-shot 3d point cloud semantic segmentation.
\newblock In \emph{Proceedings of the IEEE/CVF Conference on Computer Vision and Pattern Recognition}, pages 8873--8882, 2021.

\bibitem[Zhou and Yu(2022)]{zhou2022point}
Qiang Zhou and Chaohui Yu.
\newblock Point rcnn: An angle-free framework for rotated object detection.
\newblock \emph{Remote Sensing}, 14\penalty0 (11):\penalty0 2605, 2022.

\end{thebibliography}
}


\end{document}